\documentclass{article}

    \PassOptionsToPackage{numbers, compress}{natbib}

\usepackage[final]{neurips_2024}

\usepackage{enumitem}
\usepackage{arydshln} 
\usepackage[utf8]{inputenc} 
\usepackage[T1]{fontenc}    
\usepackage[colorlinks=true,citecolor=brown,urlcolor=gray]{hyperref}
\usepackage{url}            
\usepackage{wrapfig,lipsum,booktabs}       
\usepackage{amsfonts}       
\usepackage{nicefrac}       
\usepackage{microtype}      
\usepackage[dvipsnames]{xcolor}  

\usepackage{colortbl}
\usepackage{amssymb}
\usepackage{pifont}
\usepackage{rotating}
\usepackage{makecell}

\usepackage{bbm}

\usepackage{caption}
\usepackage{subcaption}
\usepackage{amsmath}
\usepackage{xspace}
\usepackage{multirow}

\usepackage{algorithm}
\usepackage{algpseudocode}

\usepackage[titletoc]{appendix}

\usepackage{pdflscape}

\usepackage{tabularray}

\newcommand{\cmark}{\ding{51}}%
\newcommand{\starmark}{\ding{72}}
\newcommand{\trianglemark}{\ding{116}}
\newcommand{\ours}{BLoB\xspace}
\newcommand{\loraens}{ENS\xspace}
\newcommand{\loramle}{MLE\xspace}
\newcommand{\loramap}{MAP\xspace}
\newcommand{\loramcd}{MCD\xspace}
\newcommand{\loralap}{LAP\xspace}
\newcommand{\loranaivebbb}{BBB\xspace}

\newlength\savewidth

\newcolumntype{C}{>{\centering\let\newline\\\arraybackslash\hspace{0pt}}m{2cm}}

\newcommand{\StartMenu}{\raisebox{-0.15cm}{\includegraphics[scale=0.025]{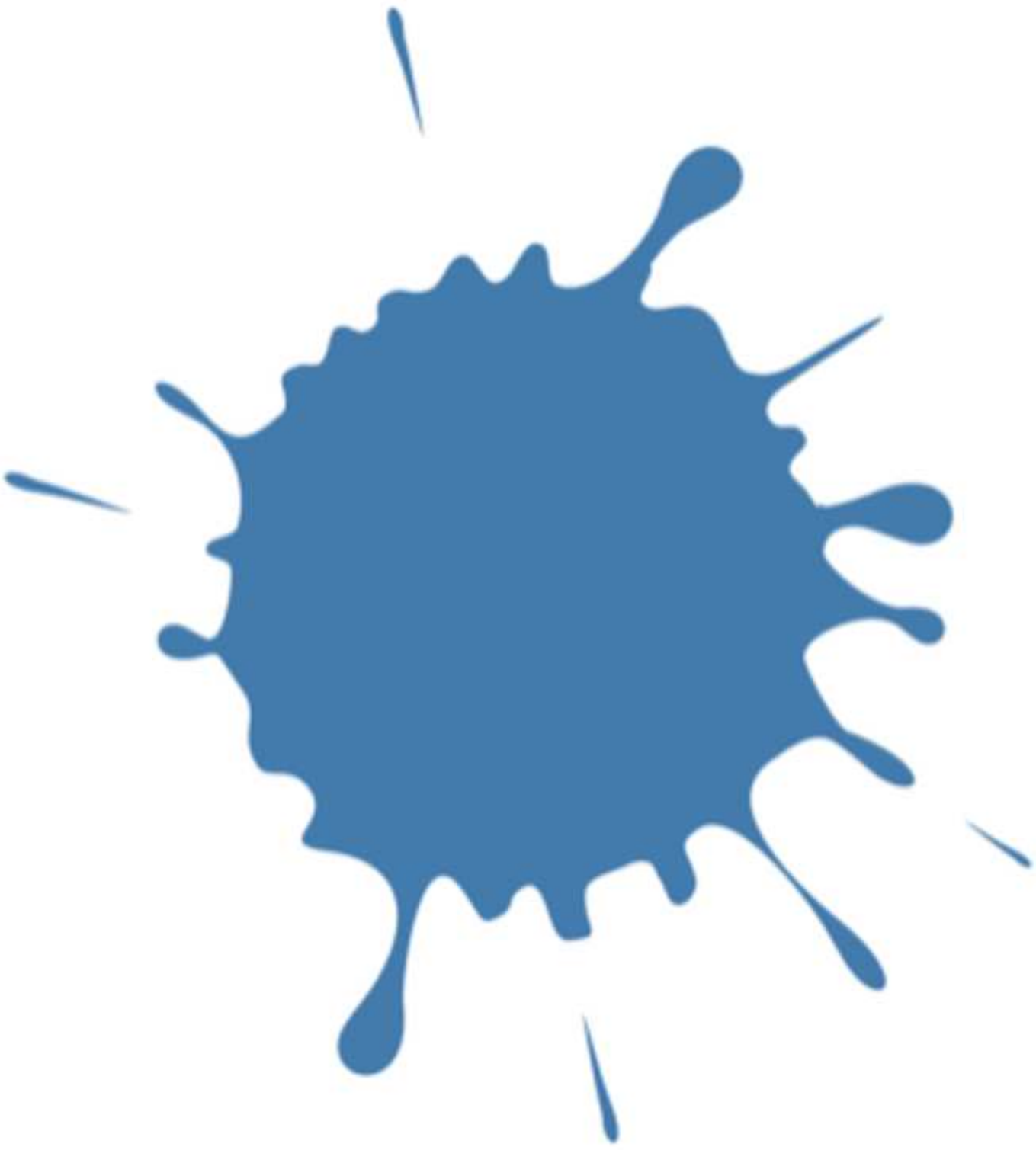}}}%
\newcommand{\SODA}[1]{%
   \StartMenu
   \foreach \x in {#1} {%
   \texttt{\textbf{\x}}%
   }%
}%

\usepackage{amsmath,amsfonts,bm,eqnarray}
\usepackage{cancel}
\usepackage{amsthm}
\usepackage{tikz}
\usetikzlibrary{tikzmark}

\newtheorem{theorem}{Theorem}[section]

\newtheorem*{remark}{Remark}

\newtheorem{innercustomgeneric}{\customgenericname}
\providecommand{\customgenericname}{}
\newcommand{\newcustomtheorem}[2]{%
  \newenvironment{#1}[1]
  {%
   \renewcommand\customgenericname{#2}%
   \renewcommand\theinnercustomgeneric{##1}%
   \innercustomgeneric
  }
  {\endinnercustomgeneric}
}

\newcustomtheorem{customThm}{Theorem}
\newcustomtheorem{customLemma}{Lemma}
\newcustomtheorem{customCor}{Corollary}
\newcustomtheorem{customProposition}{Proposition}

\newcommand{\vectorize}{\operatorname{vec}}
\newcommand{\diag}{\operatorname{diag}}

\def\Tabref#1{Table~\ref{#1}}

\def\Thmref#1{Theorem~\ref{#1}}
\def\figref#1{Fig.~\ref{#1}}
\def\Figref#1{Fig.~\ref{#1}}

\def\appref#1{Appendix~\ref{#1}}
\def\Secref#1{Sec.~\ref{#1}}

\def\eqref#1{equation~\ref{#1}}
\def\Eqref#1{Eqn.~\ref{#1}}
\def\eqnref#1{Eqn.~\ref{#1}}

\def\algref#1{algorithm~\ref{#1}}
\def\Algref#1{Algorithm~\ref{#1}}

\def\1{\bm{1}}

\def\vzero{{\bm{0}}}

\def\vmu{{\bm{\mu}}}
\def\vtheta{{\bm{\theta}}}
\def\va{{\bm{a}}}
\def\vb{{\bm{b}}}

\def\vh{{\bm{h}}}

\def\vs{{\bm{s}}}
\def\vt{{\bm{t}}}

\def\vx{{\bm{x}}}
\def\vy{{\bm{y}}}
\def\vz{{\bm{z}}}
\def\vepsilon{{\bm{\varepsilon}}}

\def\vepsilon{{\boldsymbol{\epsilon}}}

\def\vsigma{{\boldsymbol{\sigma}}}
\def\vphi{{\boldsymbol{\phi}}}
\def\vrho{{\boldsymbol{\rho}}}

\def\mA{{\bm{A}}}
\def\mB{{\bm{B}}}

\def\mD{{\bm{D}}}
\def\mE{{\bm{E}}}

\def\mG{{\bm{G}}}
\def\mH{{\bm{H}}}
\def\mI{{\bm{I}}}

\def\mM{{\bm{M}}}

\def\mO{{\bm{O}}}
\def\mP{{\bm{P}}}

\def\mR{{\bm{R}}}
\def\mS{{\bm{S}}}
\def\mT{{\bm{T}}}
\def\mU{{\bm{U}}}
\def\mV{{\bm{V}}}
\def\mW{{\bm{W}}}
\def\mX{{\bm{X}}}

\def\mZ{{\bm{Z}}}

\def\mSigma{{\bm{\Sigma}}}
\def\mOmega{{\bm{\Omega}}}

\DeclareMathAlphabet{\mathsfit}{\encodingdefault}{\sfdefault}{m}{sl}
\SetMathAlphabet{\mathsfit}{bold}{\encodingdefault}{\sfdefault}{bx}{n}

\def\gD{{\mathcal{D}}}

\def\gF{{\mathcal{F}}}

\def\gL{{\mathcal{L}}}

\def\gN{{\mathcal{N}}}

\newcommand{\E}{\mathbb{E}}

\renewcommand{\tilde}{\widetilde}
\renewcommand{\hat}{\widehat}
\renewcommand{\frac}{\tfrac}

\title{\vspace{-0.12cm}\SODA{\ours}: Bayesian Low-Rank Adaptation by Backpropagation for Large Language Models}

\author{%
  Yibin Wang \thanks{Equal Contribution. 
  $^1$Rutgers University.
  $^2$MIT-IBM Watson AI Lab.
  $^{\text{\textdagger}}$Correspondence to: Yibin Wang <yibin.wang@rutgers.edu>, Haizhou Shi <haizhou.shi@rutgers.edu>, 
  Hao Wang <hw488@cs.rutgers.edu>.}\space\space{$^{\text{\textdagger} 1}$} \\
  \And
  Haizhou Shi $^{*\text{\textdagger}1}$\\
  \And
  Ligong Han $^{12}$ \\
  \And
  Dimitris Metaxas $^{1}$ \\
  \And
  Hao Wang $^{\text{\textdagger} 1}$\\
}

\begin{document}

\maketitle

\begin{abstract}
Large Language Models~(LLMs) often suffer from overconfidence during inference, particularly when adapted to downstream domain-specific tasks with limited data. 
Previous work addresses this issue by employing approximate Bayesian estimation \emph{after} the LLMs are trained, enabling them to quantify uncertainty. 
However, such post-training approaches' performance is severely limited by the parameters learned \emph{during} training. 
In this paper, we go beyond post-training Bayesianization and propose \textbf{B}ayesian \textbf{Lo}w-Rank Adaptation by \textbf{B}ackpropagation~(\textbf{\ours}), an algorithm that continuously and jointly adjusts both the mean and covariance of LLM parameters throughout the whole fine-tuning process. 
Our empirical results verify the effectiveness of \ours in terms of generalization and uncertainty estimation, when evaluated on both in-distribution and out-of-distribution data.
{Code is available at \url{https://github.com/Wang-ML-Lab/bayesian-peft}}.
\end{abstract}

\section{Introduction}
\label{sec:intro}

Despite the recent advancements in Large Language Models~(LLMs)~\cite{biderman2023pythia,wei2022emergent,wei2021finetuned,min2022rethinking,chowdhery2023palm,anil2023palm,touvron2023llama,touvron2023llama2,radford2019language,brown2020language,achiam2023gpt,achiam2022chatgpt}, addressing the challenges of reliability and responsibility remains imperative~\cite{huang2024learn,PACE,VALC}. LLMs often produce overconfident responses detached from factual grounding, posing potential harm to users~\cite{amodei2016concrete,weidinger2021ethical,kadavath2022language,huang2023survey,tian2023just, kuhn2023semantic, azaria2023internal, yin2023large, xiong2023can, zhang2023r, gupta2024language, nikitin2024kernel, yadkori2024believe, kapoor2024large}. Therefore, accurately estimating response confidence (or uncertainty) is crucial to preemptively intervene before harm occurs. Current research predominantly focuses on eliciting the internal capability of uncertainty estimation of LLMs. For example, studies suggest that verbalized uncertainty yields better-calibrated results compared to conditional probability~\cite{tian2023just, kapoor2024large}.


While effective, the aforementioned methods do not offer a universal solution for expressing LLM uncertainty across all scenarios, especially when adapted~\cite{GRDA} to domain-specific corpora, human preferences, or downstream tasks~\cite{kadavath2022language}. Even a well-calibrated LLM may struggle to estimate uncertainty during fine-tuning due to 
catastrophic forgetting of general knowledge~\cite{shi2024continual}. Moreover, when applied to limited-scale downstream tasks, excessively over-parameterized LLMs can rapidly overfit, leading to overconfidence. Thus, enabling accurate uncertainty estimation of LLMs is vital for their reliable and responsible deployment.

Bayesian methods emerge as a natural solution for learning uncertainty estimation abilities among their counterparts~\cite{tierney1986accurate,blundell2015BBB,BDL,wang2016natural,gal2016dropout,kendall2017what,lakshminarayanan2017simple,maddox2019simple,BIN,liu2020simple,wang2020survey,daxbergeer2021laplace,wilson2022bayesian}. These methods model predictive uncertainty $P(\vy|\vx,\gD)$ by marginalizing the posterior parameter distribution $P(\vtheta|\gD)$ after observing the dataset $\gD$:
\begin{align}
    P(\vy|\vx, \gD) &= \int P(\vy|\vx, \vtheta) P(\vtheta|\gD) d \vtheta. 
\end{align}
However, adapting the Bayesian framework to LLMs poses significant challenges. LLM architectures typically incorporate complex components, including non-linear activation functions, rendering exact Bayesian inference of parameter posteriors intractable, i.e., unable to compute the integral precisely. Consequently, finding an accurate approximation algorithm for the true posterior distribution becomes a primary challenge. Additionally, modeling parameter posterior distributions demands extra memory space, imposing a prohibitive burden on systems due to the massive scale of LLMs.

Contemporary methods leverage Parameter-Efficient Fine-Tuning (PEFT) to reduce the number of tunable parameters, thus alleviating computational and storage resource burdens~\cite{ding2023parameter,hu2022lora,edalati2022krona,zhang2020side,li2021prefix,lester2021power}. Built on this, recent research explores Bayesianizing only the PEFT module during fine-tuning to calibrate LLMs~\cite{balabanov2024uncertainty,wang2023lora,yang2023bayesian,onal2024gaussian}, somewhat relieving the burden of introducing more parameters for posterior approximation. However, initial investigations suggest that straightforward combinations of PEFT and basic Bayesian techniques like Monte-Carlo Dropout (MCD, \cite{gal2016dropout}) or Deep Ensemble (ENS, \cite{lakshminarayanan2017simple,balabanov2024uncertainty,wang2023lora}) yield only marginal improvements in generalization and uncertainty estimation. The most promising results to date involve Kronecker factorized Laplace approximation, applied after maximum a posteriori (MAP) estimation provided by any optimization algorithm~\cite{yang2023bayesian}. Nevertheless, we argue that such post-training procedures bifurcate posterior approximation into two stages, inevitably leading to suboptimal estimation.


To address this challenge, we propose 
\textbf{B}ayesian \textbf{Lo}w-Rank Adaptation by \textbf{B}ackpropagation~(\textbf{\ours}), a Bayesian Deep Learning framework for fine-tuning LLMs with LoRA. 
\ours jointly estimates the low-rank variational distributions' mean and covariance throughout the entire fine-tuning stage via backpropagation. Unlike methods relying on post-training approximation, \ours enables simultaneous estimation of both the parameter mode (i.e., the mean if one assumes Gaussian distributions) and the parameter variance. Random sampling of model parameters based on variance estimation can enhance mode estimation. It thereby improves model performance in terms of accuracy and uncertainty estimation on both in-distribution and out-of-distribution datasets, as verified by our extensive experiments across multiple datasets. 
In summary, our contributions are:
\begin{itemize}
    \item We propose a principled Bayesianization framework for Low-Rank Adaptation (LoRA) in Large Language Models (LLMs) by assuming that full weights' approximate posterior distribution has a low-rank structure containing a linear combination of independent Gaussian distributions. 
    \item We show that, under mild conditions, optimization of the full-weight variational distribution can be done efficiently in the low-rank space of the weight update matrices.
    \item We introduce \ours, a variational Bayesian low-rank adaptation framework for LLMs that jointly learns the mean and covariance of the variational distribution during fine-tuning.
    \item Extensive evaluations demonstrate the superiority of \ours in terms of generalization and uncertainty estimation across different scenarios.
\end{itemize}

\section{Preliminaries}
\label{sec:preliminaries}

In this section, we describe the notation as well as some preliminaries. 



\textbf{Notation.}\quad 
In this paper, scalars are denoted by lowercase letters, vectors by lowercase boldface letters, and matrices by uppercase boldface letters. Probability, expectation, and the dataset are denoted by $P$, $\E$, and $\gD$, respectively.
We use $[m]=\{1,2,\cdots,m\}$ to denote the set of consecutive integer numbers starting from $1$ and ending at $m$.
For a matrix $\mX=[\vx_1, \cdots, \vx_n]\in \mathbb{R}^{m\times n}$, we use $\vectorize(\mX)= 
[\vx_1^\top, \vx_2^\top, \cdots, \vx_n^\top]^\top
\in \mathbb{R}^{(mn)\times 1}$ to denote the vectorization operation; 
we use $\|\mX\|_p=\left[\sum_{ij}|X_{ij}|^p\right]^{\nicefrac{1}{p}}$ to define the $p$-norm of a matrix. 
We use $\otimes$ and $\circ$ to denote the Kronecker product and the element-wise product, respectively. 

\subsection{Low-Rank Adaptation (LoRA)}
\label{sec:preliminaries-lora}

Inspired by the pioneering work on identifying and leveraging the low intrinsic rank of over-parameterized models during fine-tuning~\cite{li2018measuring,aghajanyan2021intrinsic}, Low-Rank Adaptation (LoRA) assumes a low rank for the network's weight updates~\cite{hu2022lora}. Typically in a single linear layer, LoRA decomposes each update matrix $\Delta \mW = \mB\mA$ into the product of two low-rank matrices, where $\mB \in \mathbb{R}^{m \times r}$ and $\mA \in \mathbb{R}^{r \times n}$. Here, $m$, $n$, and $r$ denote the number of input neurons, output neurons, and the rank of the decomposition, respectively~\cite{hu2022lora}. The forward pass of the linear layer with LoRA is formulated as:
\begin{align}
    \vz &= \mW_0 \vh + \Delta \mW \vh = \mW_0 \vh + \mB\mA \vh,
\end{align}
where $\vh$ and $\vz$ denote the input and output of the layer.
Since the rank $r \ll \min\{m, n\}$ is significantly smaller than the numbers of input and output neurons (e.g., $r = 8 \ll m = n= 4096$ in the attention layer~\cite{hu2022lora}), LoRA can drastically reduce the number of trainable parameters by approximately three orders of magnitude compared to full-parameter fine-tuning, while achieving comparable performance to the full-rank fine-tuning. 
This also leads to a similar reduction in memory consumption for storing optimizer states, thereby reducing the hardware requirements for fine-tuning LLMs to a great extent. 

\subsection{Variational Bayesian Networks~(VBNs)}
\label{sec:preliminaries-bbb}
Bayesian Neural Networks (BNNs) estimate the posterior distributions of network parameters rather than relying on single-point estimates~\cite{bishop2006pattern, wang2020survey}. Due to the intractability of exact inference of the true posterior, Variational Bayesian Networks (VBNs) approximate the true posterior using a variational distribution; this is done by minimizing its KL divergence from the true posterior distribution~\cite{hinton1993keeping, graves2011practical, blundell2015BBB}. 
Specifically, if the weights $\mW$'s variational distribution $q(\mW|\vtheta)$ is parameterized by $\vtheta$, minimizing the divergence $\operatorname{KL}[q(\mW|\vtheta) \| P(\mW|\gD)]$ is equivalent to minimizing the following variational free energy with respect to $\vtheta$~\cite{neal1998view, yedidia2000generalized, friston2007variational}:
\begin{align}
\label{eq:vfe}
\gF(\gD, \vtheta) \triangleq - \mathbb{E}_{q(\mW|\vtheta)}[\log P(\gD|\mW)] + \operatorname{KL}[q(\mW|\vtheta) \parallel P(\mW)].
\end{align}
The final formulation of the objective function in \Eqref{eq:vfe} offers another interpretation beyond minimizing the KL divergence between the variational and true posterior distributions~\cite{blundell2015BBB}. 
Specifically, the first term maximizes the likelihood of the data, while the second term regularizes the variational distribution $q(\mW|\vtheta)$. 
We refer to the first term as the likelihood cost and the second term as the complexity cost.
Optimizing these two terms involves balancing the expressiveness of the approximate posterior distribution and its simplicity. 

Optimizing the first term of \Eqref{eq:vfe} requires integrating out the parameterized variational distribution, necessitating Monte Carlo gradient estimation~\cite{lecun1985procedure, rumelhart1986learning}. Using this approach, we can incorporate the re-parameterization trick to enable backpropagation of the gradient to the underlying parameter $\vtheta$~\cite{opper2009variational, Kingma2014, rezende2014stochastic}. In Bayes By Backprop (BBB)~\cite{blundell2015BBB}, the variational distribution is further simplified as a diagonal Gaussian $\gN(\vmu, \vsigma^2)$, where $\vsigma=\log(1+\exp(\vrho))$ ensures the standard deviation is positive. Then we have the Monte-Carlo estimation of \Eqref{eq:vfe} that can pass the gradient to $\vtheta$:
\begin{align}
    \gF(\gD, \vtheta) &\approx 
    - \frac{1}{K} \sum_{k=1}^{K} \log P(\gD|\mW_k) + \frac{1}{K} \sum_{k=1}^{K} [\log  q(\mW_k|\vtheta) - \log P(\mW_k)] ,
\end{align}
where $\mW_k = \vmu + \log(1+\exp(\vrho)) \odot \vepsilon_k$ is the $k$-th sample of the weights yielded by parameterization and $\vepsilon_k \sim \gN(\vzero, \mI)$. 
In BBB, the authors assume the prior distribution $P(\mW) = \pi \gN(\vzero, \vsigma_1^2) + (1 - \pi) \gN(\vzero, \vsigma_2^2) $ to be a mixture of Gaussians. Consequently, they optimize the second term based on weight sampling. In different scenarios, a simpler form of the prior, which allows for a closed-form solution, can also be considered.
Although our proposed method is largely based on the existing framework of BBB, trivially combining BBB with LoRA does not yield satisfactory results. It is important to note that our specific designs are necessary to encourage the fast convergence of the variational distribution, which will be introduced later in \Secref{sec:method}.



\section{Methodology}
\label{sec:method}
In this section, we formally introduce our proposed method, \textbf{B}ayesian \textbf{Lo}w-Rank Adaptation by \textbf{B}ackpropagation (\ours). 
We begin by discussing the design choices for Bayesianizing LoRA parameters in \Secref{sec:method-posterior}, highlighting the assumptions \ours makes about the approximate posterior in the full-weight space. 
Next, in \Secref{sec:method-prior}, we explore the low-rank structure of the prior distribution in the full-weight space, which in turn motivates our choice of prior distributions in the low-rank parameter space. 
In \Secref{sec:method-reparam}, we introduce our parameterization method for the variational distributions. In \Secref{sec:method-flipout}, we integrate Flipout~\cite{wen2018flipout} into LoRA for improved sampling efficiency and faster convergence. 
Finally, we present the complete algorithmic description of \ours in \Secref{sec:method-final-algo}. Proof of the theorems and claims in this section can be found in \appref{app:proof}.


\begin{figure}[h] 
\centering 
\includegraphics[width=1\textwidth]{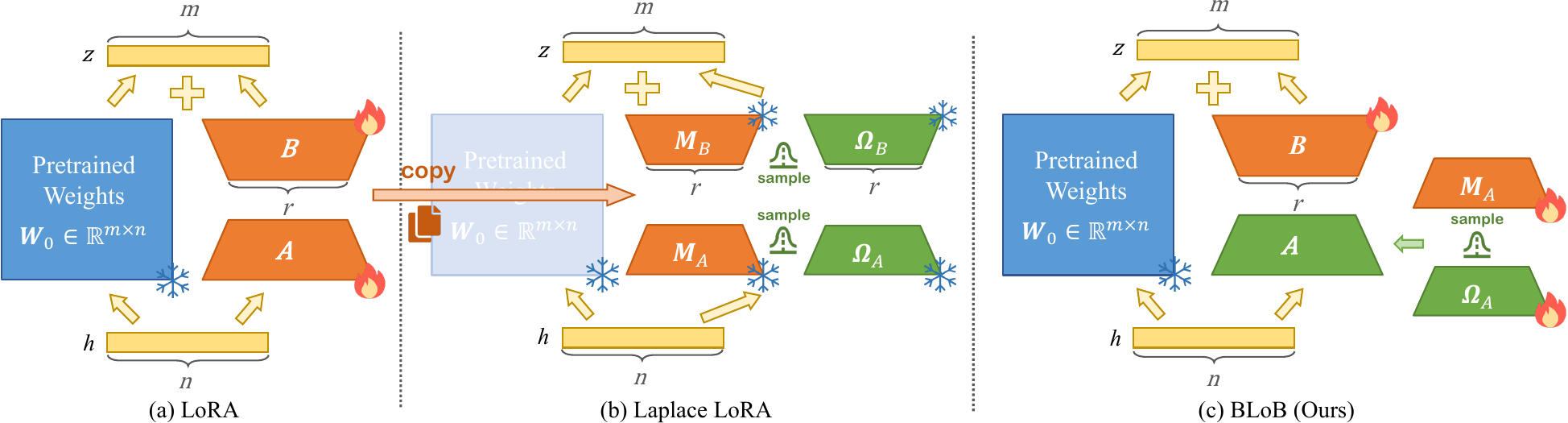} 
\caption{
Overview of our Bayesian Low-Rank Adaptation by Backpropagation, i.e., \ours~\textbf{(right)} as well as comparison with existing methods such as LoRA \textbf{(left)} and Laplace LoRA \textbf{(middle)}. 
} 
\label{fig:ours} 
\end{figure}

\subsection{Low-Rank Variational Approximate Posterior Distribution: LoRA Bayesianization}
\label{sec:method-posterior}
\textbf{Asymmetric LoRA Bayesianization.} 
In LoRA~\cite{hu2022lora}, the weights are treated asymmetrically. $\mA$ is randomly initialized, usually from the standard normal distribution or using Kaiming initialization, while $\mB=\vzero$ is initialized as a zero matrix to 
ensure that the model fully retains the capabilities of the pre-trained weights at the start of fine-tuning. 
The trivial solution of estimating the variational approximate posterior for the entire set of LoRA parameters can significantly hinder training convergence. For example, consider the Gaussian posteriors $q(\mA|\vtheta) = \gN(\mA|\mM_A, \mOmega_A^2)$ and $q(\mB|\vtheta) = \gN(\mB|\vzero, \mOmega_B^2)$, where $\mOmega_A$ and $\mOmega_B$ are variance estimates added to $\mA$ and $\mB$, respectively. 
Although the expectation $\E_{\mA, \mB}[(\mW_0 + \mB\mA)\vx] = \mW_0\vx + \E_{\mA, \mB}[\mB\mA\vx] = \mW_0\vx$ preserves the functionality of the pre-trained model, accurate estimation requires an impractically large number of weight samples. 
Such variational distributions lead to significant fluctuations during the early stages of fine-tuning, unless the initial variance of $\mB$, $\mOmega_B\rightarrow \vzero^+$, is intentionally minimized towards zero. 
Therefore, 
we take an \emph{asymmetric} approach to initialize $\mOmega_B = \vzero$ and keep it fixed throughout the fine-tuning process. This, in effect, gives up Bayesian modeling of the $\mB$ component and focuses only on the posterior of $\mA$ in LoRA, as shown in~\Figref{fig:ours}.


\textbf{Additional Advantages.}
In addition to reducing sampling noise and improving convergence speed, our Bayesianization design has two further advantages. First, compared to modeling the variational distributions of both $\mA$ and $\mB$, our approach significantly reduces additional memory cost by approximately 50\% per layer.
Second, our design is equivalent to finding a posterior estimate for the full-weight matrix with a low-rank structure. 
For instance, by assuming a deterministic $\mB$ and Bayesianizing $P(\mA|\vtheta) = \gN(\mA|\mM, \mOmega^2)$, each element of the full weight matrix $W_{ij}$ is calculated as
\begin{align}\label{eq:bayesianization}
    W_{ij} &= W_{0,ij} + \sum_{k=1}^r B_{ik} A_{kj},
\end{align}
where $A_{kj} \sim \gN(M_{kj}, \Omega_{kj}^2)$ is drawn independently $\forall k\in [r]$.
It is noteworthy that due to the low-rank structure defined in \Eqref{eq:bayesianization}, the full-weight parameters of $\mW$ are no longer independent from each other. The correlation among them can be reflected by the following theorem:

\begin{theorem}[\textbf{Variational Distribution of the Full-Weight Matrix in \ours}]\label{thm:posterior}
    With the pre-trained weight matrix $\mW_0 \in\mathbb{R}^{m\times n}$ and the low-rank weight update matrix $\mB \in \mathbb{R}^{m\times r}$, suppose that the variational distribution of the other low-rank update matrix $\mA \in \mathbb{R}^{r\times n}$ is Gaussian with $q(\mA|\vtheta=\{\mM, \mOmega\})=\prod_{ij} \gN (A_{ij}|M_{ij},\Omega_{ij}^2)$, where $\mM=[M_{ij}] \in \mathbb{R}^{r\times n}$ and $\mOmega=[\Omega_{ij}] \in \mathbb{R}^{r\times n}$ are its mean and standard deviation, respectively. The equivalent variational distribution defined on the full weight matrix $\mW$ as in \Eqref{eq:vfe} is given by
    \begin{align}
        q(\vectorize(\mW)|\mB, \vtheta) &= \gN(\vectorize(\mW)|\vmu_q, \mSigma_q), \label{eq:qW}\\
        \text{where }\quad 
        \vmu_q &= \vectorize(\mW_0+\mB\mM), \\
        \mSigma_q &= [\mI_n \otimes \mB] [\diag(\vectorize(\mOmega)^2)] [\mI_n \otimes \mB^\top].  \label{eq:Sigma_q}
    \end{align}
\end{theorem}
\Thmref{thm:posterior} shows that our asymmetric LoRA Bayesianization is equivalent to using a Gaussian variational distribution for the full weight $\mW$ (i.e., \eqnref{eq:qW}), with a flexible covariance matrix (i.e., \eqnref{eq:Sigma_q}), to approximate the posterior distribution of the full weight $\mW$. 


\begin{remark}
    The covariance matrix $\mSigma_q$ is strictly singular, which consequently inspires us to design a prior $P(\mW)$ with such low-rank structure in \Secref{sec:method-prior}.
    Previous work on low-rank Gaussians typically considers covariance with a similar structure $\mD^2+\mSigma_q$, where $\mD$ is diagonal~\cite{titsias2014doubly,ong2018gaussian,tan2018gaussian,tomczak2020efficient,park2024density}. However, sampling from a Gaussian with this structure requires sampling noise of the same shape as the full-weight matrix, which is not parameter-efficient; we therefore do not adopt this in our work.
\end{remark}

\subsection{Low-Rank Prior Distribution}\label{sec:method-prior}
In \Eqref{eq:vfe}, optimizing the KL divergence between the variational and prior distributions in the space of full weights can be burdensome. Therefore, 
we assume the prior distribution of the full weights to be a low-rank Gaussian, with its mean centered at the pre-trained weights $\vectorize(\mW_0)$ and its covariance matrix parameterized by a rank-$r^\prime$ matrix $\tilde{\mR} \in \mathbb{R}^{(mn) \times r^\prime}$:
\begin{align}\label{eq:prior}
    \nonumber 
    P(\vectorize(\mW)) &= \gN(\vectorize(\mW)|\vmu_p, \mSigma_p), \\
    \text{where }\quad 
    \vmu_p &= \vectorize(\mW_0), \\
    \nonumber 
    \mSigma_p &= \tilde{\mR}\tilde{\mR}^\top.
\end{align}
Assuming a low-rank prior distribution and designing an appropriate $\tilde{\mR}$ allows us to optimize the KL divergence in the decomposed low-rank weight space, as suggested by the following theorem.

\begin{theorem}[\textbf{Efficient Computation of Full-Weight KL Divergence}]\label{thm:prior}
    Suppose the pre-trained weights $\mW_0$, update matrix $\mB$, and the variational distribution $q(\mA|\vtheta)$ are defined as in \Thmref{thm:posterior}, 
    and the prior distribution of the full-weight matrix $P(\vectorize(\mW))$ is defined as \Eqref{eq:prior}. 
    Consider the Gaussian prior distribution 
    $P(\mA)=\prod_{ij} \gN (A_{ij}|0,\sigma_p^2)$; we then have:
    \begin{align}
        \operatorname{KL}[ q(\vectorize(\mW)|\mB,\vtheta) \| P(\vectorize(\mW))]
        &= \operatorname{KL}[ q(\mA|\vtheta) \| P(\mA)],
    \end{align}
    if $\tilde{\mR}=[\sigma_p \mI_n \otimes \mR]$, where $\mR$ satisfies $\mR\mR^\top = \mB\mB^\top$.
\end{theorem}


\Thmref{thm:prior} shows that with a proper $\tilde{\mR}$, one can compute the KL divergence for the high-dimensional full weight $\vectorize(\mW)$ simply by computing the KL divergence for $\mA$, which is much lower-dimension, more parameter-efficient, more memory-efficient, and faster. 
{Note that the Gaussian distributions we define for both the prior and the posterior are degenerate. However, they are valid for probabilistic inference~\cite{schoeman2022degenerate}, as (i)~their probability density is well-defined, and (ii)~their KL divergence is computable under the assumptions of \Thmref{thm:prior}. See \appref{app:proof-main} for a detailed discussion.}

Concretely, we assume that the prior distribution in \ours follows the low-rank structure described in \Thmref{thm:prior} and minimize the KL divergence term for the low-rank component $\mA$ using its analytical solution in \Eqref{eq:vfe}:
\begin{align}\label{eq:closed costf}
    \operatorname{KL}[ q(\mA|\vtheta=\{\mM,\mOmega\}) \| P(\mA) ]
    &=  
    \frac{1}{2\sigma_p^2}(\|\mM\|_2^2+\|\mOmega\|_2^2) - \sum_{ij}\log\Omega_{ij}.
\end{align}


\subsection{Parameterization of the Low-Rank Variational Distribution}
\label{sec:method-reparam}
The parameterization of the Gaussian variational distribution 
$q(\mA|\vtheta)$ significantly affects the convergence speed of the KL term in \eqnref{eq:closed costf}. 
The mean matrix $\mM$ of $q(\mA|\vtheta)$ has no additional constraints, we therefore parameterize it directly as the output of a neural network. 
Each entry of $q(\mA|\vtheta)$'s diagonal covariance matrix $\mOmega$ (i.e., standard deviation) is non-negative; we therefore use
element-wise parameterization $\Omega_{ij} = G_{ij}^2$, where $\mG = [G_{ij}] \in \mathbb{R}^{r \times n}$ is the real parameter matrix that determines the standard deviation $\mOmega$. Since $\mOmega$ is usually initialized with small positive values close to zero, our parameterization method provides large gradients initially, contributing to the rapid decrease of the KL term. We further show, both theoretically and empirically, that our parameterization method, unlike BBB's softplus function $\log(1+\exp(\cdot))$, is crucial for the fast convergence of $\mOmega$ when $q(\mA|\vtheta)$ is close to the prior distribution $P(\mA)$ (see more analysis in Appendix \ref{app:reparameter}).

\subsection{On Improving the Sample Efficiency of \ours}
\label{sec:method-flipout}
\textbf{Improving Sample Efficiency with Flipout.} 
One main challenge in estimating the variational distribution (i.e., the approximate posterior) during fine-tuning lies in the sample efficiency of the weights~\cite{wiese2023towards,dusenberry2020efficient,liu2020simple}. 
During mini-batch stochastic gradient descent, a batch of examples typically share the same weights drawn from the variational distribution. This can lead to slow convergence of the likelihood cost in \Eqref{eq:vfe}.
Drawing inspiration from \cite{wen2018flipout}, we introduce the technique of flipout to speed up the sampling procedure of our low-rank variational distributions $q(\mA|\vtheta)$. 

\begin{algorithm}[t]
\caption{\quad\textbf{B}ayesian \textbf{Lo}w-Rank Adaptation by \textbf{B}ackpropagation (\textbf{\ours})}\label{alg:blob}
\begin{algorithmic}[1]
\Require dataset $\gD$, pre-trained weight $\mW_0$, low-rank component $\mB$, $\vtheta =\{\mM, \mG\}$ for parameterizing the mean and variance of $\mA$;
\Require prior standard deviation $\sigma_p$, initialization hyperparameter $\epsilon$, number of input features $n$;
\Require number of samples during training $K$, number of iterations $T$, learning rate $\eta$;
\vspace{0.5em}

\State $\mG \sim \mathcal{U}(\frac{\epsilon}{\sqrt{2}}, \epsilon)$, $\mM \sim \mathcal{U}\left(-\sqrt{\frac{6}{n}}, \sqrt{\frac{6}{n}}\right)$ \Comment{Initialization of $\mA$'s parameters.}
\State $\mB \gets \mathbf{0} $  \Comment{Initialization of $\mB$.}

\For{$t = 1, \cdots, T$}
    \State Sample a mini-batch of data $\gD_t\sim\gD$ {containing $b$ samples}.
    \For{$k = 1, \cdots, K$}
        \State Sample batched noise $\mE_k \sim \mathcal{N}\left(\vzero, \mI\right)$. \Comment{Sample the noise.}
        \State {Let $\{\tilde{\mE}_{kj}\}_{j=1}^{b}\leftarrow\operatorname{BLoBFlipout}(\mE_k)$.}\Comment{\eqnref{eq:flipout}}
        \State {Let $\{\mA_{kj} = \mM + \mG^2 \circ \tilde{\mE}_{kj}\}_{j=1}^{b}$.}
    \EndFor
    \State {Let $\hat{\gF}_t =
    - \frac{1}{Kb}\sum_{k=1}^K\sum_{j=1}^b\log P(\gD_t|\mA_{kj},\mB) + \frac{1}{2\sigma_p^2}(\|\mM\|_2^2+\|\mG\|_2^4) - 2\sum_{ij}\log G_{ij}$.}\\
    \Comment{\eqnref{eq:final elbo} and  \ref{eq:closed costf}.}
    \State Calculate the gradient w.r.t. the parameters: \par 
    \quad $\Delta_{\mM} = \nicefrac{\partial \hat{\gF}_t}{\partial \mM}$, $\Delta_{\mG} = \nicefrac{\partial \hat{\gF}_t}{\partial \mG}$, $\Delta_{\mB} = \nicefrac{\partial \hat{\gF}_t}{\partial \mB}$.
    \State Update the parameters: \par
        \quad $\mM \gets \mM - \eta \Delta_{\mM}$; \par
        \quad $\mG \gets \mG - \eta \Delta_{\mG}$; \par
        \quad $\mB \gets \mB - \eta \Delta_{\mB}$ .
\EndFor
\end{algorithmic}
\end{algorithm}

\textbf{LoRA Flipout.}
Unlike the original approach, which applies rank-1 random flipping to the full weights, we apply flipout exclusively to the low-rank component $\mA$.
Specifically, suppose we have a mini-batch of input vectors $\mH\in \mathbb{R}^{n\times b}$, where $b$ represents the batch size. We randomly sample two low-rank flipping matrices $\mS\in \{-1,+1\}^{n\times b}$ and $\mT\in \{-1,+1\}^{b\times r}$. Denoting as $\mE\in \mathbb{R}^{r\times n}$ the weight noise sampled for this mini-batch, the batched output $\mZ$ after applying flipout is then
\begin{align}\label{eq:flipout}
    \mZ &= \mW_0 \mH + \mB (\mM \mH + \left[ (\mE \circ \mOmega) (\mH \circ \mS) \right] \circ \mT ),
\end{align}
It is crucial that the independent noises added to the low-rank weight noise $\Delta\mA\triangleq\mE\circ\mOmega$ ensure \emph{sampling independence across examples} within a mini-batch, thereby enhancing the sampling efficiency of the algorithm. This is done without violating the assumptions outlined in \Thmref{thm:posterior} and \ref{thm:prior}. 
{As illustrated in \algref{alg:blob}, we use $\tilde{\mE}_{kj}$ to represent the equivalent noise applied to parameter $\mA$ for the $j$-th example in the $k$-th batch after BLoBFlipout.}
Due to the low-rank structure of our Bayesianization method, the computational overhead of employing flipout in \ours is also minimal.

\subsection{\ours: Final Algorithm}
\label{sec:method-final-algo}
We are now ready to present our full \ours algorithm. 

{\textbf{During training},} under the assumptions outlined in \Thmref{thm:posterior} and \ref{thm:prior}, optimizing the evidence lower bound on the full weight $\mW$ can be efficiently done in the low-rank space, using the following final objective function:
\begin{align}
\label{eq:final elbo}
    \nonumber \gF(\gD, \mB, \vtheta) 
    &= 
    - \mathbb{E}_{q(\mW|\mB,\vtheta)}[\log P(\gD|\mW)] + \operatorname{KL}[q(\mW|\mB,\vtheta) \parallel P(\mW)]\\
    &= 
    - \mathbb{E}_{q(\mA|\vtheta)}[\log P(\gD|\mA,\mB)] + \operatorname{KL}[q(\mA|\vtheta) \parallel P(\mA)], 
\end{align}
where $\vtheta=\{\mM,\mOmega\}$ denotes the set of the parameters underlying the variational distribution of the low-rank matrix $\mA$. 
{Additionally, to trade off between data fitting and posterior approximation, we employ a KL re-weighting scheme, which is detailed in \appref{kl-reweighting}.} 
The full algorithmic description of \ours training is shown in \Algref{alg:blob}.

\textbf{During inference, }for an input $\vx$, we approximate the expected output distribution $P(\vy|\vx)$ of \ours by drawing $N$ samples from the variational distribution $q(\mW|\vtheta)$. Empirically, $N=10$ provides a good balance between estimation quality and computational efficiency:
\begin{align}
    \E_{q(\mW|\vtheta)}[P(\vy|\vx,\mW)] 
    &\approx \frac{1}{N} \sum_{n=1}^{N} P(\vy|\vx,\mW_n), \quad \mW_n \sim q(\mW|\vtheta).
\end{align}

\section{Experiments}
\label{sec:experiments}
In this section, we compare our \ours with existing methods on real-world datasets. 
\Secref{sec:experiments-setting} introduces the experimental settings, including baselines, fine-tuning, and evaluation protocols. We then evaluate \ours's generalization and uncertainty estimation abilities in both in-distribution (\Secref{sec:exp-id}) and out-of-distribution scenarios (\Secref{sec:exp-ood}). 


\subsection{Settings}
\label{sec:experiments-setting}

\textbf{Fine-tuning and Evaluation.}
We implement \ours in the PEFT library~\cite{peft} and fine-tune the LlaMA2-7B~\cite{touvron2023llama2} model on common-sense reasoning tasks. Following Laplace-LoRA~\cite{yang2023bayesian}, we apply LoRA to the output layer as well as the queries and values of all the attention layers. For hyperparameters, we strictly adhere to the default settings in the PEFT library and the original LoRA paper~\cite{peft, hu2022lora} to ensure maximal reproducibility. This includes the number of training steps, learning rate, and LoRA rank $r$ (see \appref{app:implementation-ours} for details).
For common-sense reasoning tasks, we select the next token logits corresponding to possible answers from each dataset and fine-tune the LLM to maximize the likelihood of the correct token. For evaluation, in addition to Accuracy (\textbf{ACC}), we use Expected Calibration Error (\textbf{ECE}~\cite{naeini2015obtaining}) and Negative Log-Likelihood (\textbf{NLL}) to assess the models' uncertainty estimation ability (see \appref{app:implementation-metrics} for details).


\textbf{Baselines and Implementation Details.}
We compare \textbf{\ours} with state-of-the-art uncertainty estimation methods applied to the LoRA adapters of LLMs, including Monte-Carlo Dropout (\textbf{MCD})~\cite{gal2016dropout}, Bayes By Backprop (\textbf{BBB})~\cite{blundell2015BBB}, Deep Ensemble (\textbf{ENS})~\cite{lakshminarayanan2017simple,balabanov2024uncertainty,wang2023lora}, and the latest Laplace-LoRA~(\textbf{LAP})~\cite{yang2023bayesian}. 
{We also report the performance of two standard PEFT baseline methods for reference: Maximum Likelihood Estimation~(\textbf{MLE})~\cite{hu2022lora} and Maximum A Posteriori~(\textbf{MAP}).}

For MLE, we use the LoRA implementation. For MAP, we use a weight decay rate of $1e-5$. For MCD, we use an ensemble of 10 LoRAs with a dropout rate of $p=0.1$. For ENS, we independently fine-tune 3 LoRAs and average their logits during evaluation. For BBB, we adopt the default settings from the Bayesian-Torch library~\cite{krishnan2022bayesiantorch} and only Bayesianize the $\mA$ matrix, similar to \ours. 
We sample $N=10$ times for BBB during test.
We re-implement LAP and apply it to the MAP checkpoints.
We keep all \ours-specific hyperparameters consistent across all datasets. 
Typically, we set the number of samples $K=1$ during training for all our \ours experiments, which highlights \ours's sampling efficiency.
As shown in \Tabref{tab:main-llama}, we also report \ours's performance with different numbers of samples during Bayesian inference, where $N=0$ indicates directly using the mean of the weight distribution for prediction.

\begin{table*}[t]
\caption{
    \textbf{Performance of different methods applied to LoRA on Llama2-7B pre-trained weights,} {where Accuracy~(\textbf{ACC}) and Expected Calibration Error~(\textbf{ECE}) are reported in percentages.} The evaluation is done across six common-sense reasoning tasks with a shared hyper-parameter setting after 5,000 gradient steps.
    We use $N$ to represent the number of samples during inference in \ours. 
    ``$\uparrow$'' and ``$\downarrow$'' indicate that higher and lower values are preferred, respectively. 
    \textbf{Boldface} and \underline{underlining} denote the best and the second-best performance, respectively. 
}
\begin{center}
\resizebox{1\linewidth}{!}{%
\setlength{\tabcolsep}{8pt}

\begin{tabular}{clccc ccc}
	\toprule[0.12em]
	\multirow{2}{*}[-0.25em]{\textbf{Metric}} & \multirow{2}{*}[-0.25em]{\textbf{Method}} & \multicolumn{6}{c}{\textbf{Datasets}}
     \\
     \cmidrule{3-8}
     & & WG-S~\cite{wg}
     & ARC-C~\cite{arc} 
     & ARC-E~\cite{arc} 
     & WG-M~\cite{wg} 
     & OBQA~\cite{obqa} 
     & BoolQ~\cite{boolq} \\
     \midrule

     \multirow{9}{*}{ACC~($\uparrow$)} & \loramle & 68.99\scriptsize{$\pm$0.58} & \underline{69.10\scriptsize{$\pm$2.84}} & 85.65\scriptsize{$\pm$0.92} & 74.53\scriptsize{$\pm$0.66} & 81.52\scriptsize{$\pm$0.25} & 86.53\scriptsize{$\pm$0.28} \\
     
     & \loramap & 68.62\scriptsize{$\pm$0.71} & 67.59\scriptsize{$\pm$0.40} & \underline{86.55\scriptsize{$\pm$0.55}} & \underline{75.61\scriptsize{$\pm$0.71}} & 81.38\scriptsize{$\pm$0.65} & 86.50\scriptsize{$\pm$0.41} \\
     
     & \loramcd~\cite{gal2016dropout} & 
     69.46\scriptsize{$\pm$0.62} & 68.69\scriptsize{$\pm$1.30} & 86.21\scriptsize{$\pm$0.46} & \textbf{76.45\scriptsize{$\pm$0.04}} & 81.72\scriptsize{$\pm$0.10} & 87.29\scriptsize{$\pm$0.13} \\
     
     & \loraens~\cite{lakshminarayanan2017simple,balabanov2024uncertainty,wang2023lora} & \underline{69.57\scriptsize{$\pm$0.66}} & 66.20\scriptsize{$\pm$2.01} & 84.40\scriptsize{$\pm$0.81} & 75.32\scriptsize{$\pm$0.21} & 81.38\scriptsize{$\pm$0.91} & 87.09\scriptsize{$\pm$0.11} \\

     & \loranaivebbb~\cite{blundell2015BBB} & 
     56.54\scriptsize{$\pm$7.87} & 
     68.13\scriptsize{$\pm$1.27} & 
     85.86\scriptsize{$\pm$0.74} & 
     73.63\scriptsize{$\pm$2.44} & 
     82.06\scriptsize{$\pm$0.59} & 
     \textbf{87.21\scriptsize{$\pm$0.22}}
     \\
     
     & \loralap~\cite{yang2023bayesian} & 69.20\scriptsize{$\pm$1.50} & 66.78\scriptsize{$\pm$0.69}\footnotemark[1] & 80.05\scriptsize{$\pm$0.22} & 75.55\scriptsize{$\pm$0.36} & \underline{82.12\scriptsize{$\pm$0.67}} & 86.95\scriptsize{$\pm$0.09} \\

    \cmidrule{2-8}
    
     & \ours~(N=0) & 
     \textbf{70.89\scriptsize{$\pm$0.82}} & 
     \textbf{70.83\scriptsize{$\pm$1.57}} & 
     \textbf{86.68\scriptsize{$\pm$0.60}} &
     74.55\scriptsize{$\pm$1.94} & 
     \textbf{82.73\scriptsize{$\pm$0.41}} &
     86.80\scriptsize{$\pm$0.23}
     \\

     & \ours~(N=5) & 
     66.30\scriptsize{$\pm$0.62} & 
     67.34\scriptsize{$\pm$1.15} & 
     84.74\scriptsize{$\pm$0.33} &
     72.89\scriptsize{$\pm$1.25} & 
     81.79\scriptsize{$\pm$0.94} & 
     86.47\scriptsize{$\pm$0.15} 
     \\
     
     & \ours~(N=10) & 
     69.07\scriptsize{$\pm$0.34} & 
     68.81\scriptsize{$\pm$1.09} & 
     85.56\scriptsize{$\pm$0.35} & 
     73.69\scriptsize{$\pm$0.17} & 
     81.52\scriptsize{$\pm$0.74} & 
     \underline{86.99\scriptsize{$\pm$0.24}} 
     \\
     \midrule
    
     \multirow{9}{*}{ECE~($\downarrow$)} & \loramle & 29.83\scriptsize{$\pm$0.58} & 29.00\scriptsize{$\pm$1.97} & 13.12\scriptsize{$\pm$1.39} & 20.62\scriptsize{$\pm$0.74} & 12.55\scriptsize{$\pm$0.46} & 3.18\scriptsize{$\pm$0.09} \\
     
     & \loramap & 29.76\scriptsize{$\pm$0.87} & 29.42\scriptsize{$\pm$0.68} & 12.07\scriptsize{$\pm$0.55} & 23.07\scriptsize{$\pm$0.14} & 13.26\scriptsize{$\pm$0.82} & 3.16\scriptsize{$\pm$0.23} \\
     
     & \loramcd~\cite{gal2016dropout} & 
     27.98\scriptsize{$\pm$0.44} & 27.53\scriptsize{$\pm$0.80} & 12.20\scriptsize{$\pm$0.56} & 19.55\scriptsize{$\pm$0.47} & 13.10\scriptsize{$\pm$0.11} & 3.46\scriptsize{$\pm$0.16} \\
     
     & \loraens~\cite{lakshminarayanan2017simple,balabanov2024uncertainty,wang2023lora} & 28.52\scriptsize{$\pm$0.55} & 29.16\scriptsize{$\pm$2.37} & 12.57\scriptsize{$\pm$0.58} & 20.86\scriptsize{$\pm$0.43} & 15.34\scriptsize{$\pm$0.27} & 9.61\scriptsize{$\pm$0.24} \\

     & \loranaivebbb~\cite{blundell2015BBB} & 
     21.81\scriptsize{$\pm$12.95} & 
     26.23\scriptsize{$\pm$1.47} & 
     12.28\scriptsize{$\pm$0.58} & 
     15.76\scriptsize{$\pm$4.71} & 
     11.38\scriptsize{$\pm$1.07} & 
     3.74\scriptsize{$\pm$0.10}
     \\
     
     & \loralap~\cite{yang2023bayesian} & \textbf{4.15\scriptsize{$\pm$1.12}} & 16.25\scriptsize{$\pm$2.61}\footnotemark[1] & 33.29\scriptsize{$\pm$0.57} & 7.40\scriptsize{$\pm$0.27} & 8.70\scriptsize{$\pm$1.77} & \textbf{1.30\scriptsize{$\pm$0.33}} \\

    \cmidrule{2-8}

     & \ours~(N=0) & 
     20.62\scriptsize{$\pm$0.83} & 
     20.61\scriptsize{$\pm$1.16} & 
     9.43\scriptsize{$\pm$0.38} &
     11.23\scriptsize{$\pm$0.69} & 
     8.36\scriptsize{$\pm$0.38} & 2.46\scriptsize{$\pm$0.07} 
     \\

     & \ours~(N=5) & 
     10.89\scriptsize{$\pm$0.83} & 
     \underline{11.22\scriptsize{$\pm$0.35}} & 
     \underline{6.16\scriptsize{$\pm$0.23}} &
     \underline{4.51\scriptsize{$\pm$0.35}} & 
     \textbf{3.40\scriptsize{$\pm$0.57}} & 1.63\scriptsize{$\pm$0.35} 
     \\
     
     & \ours~(N=10) & 
     \underline{9.35\scriptsize{$\pm$1.37}} & 
     \textbf{9.59\scriptsize{$\pm$1.88}} & 
     \textbf{3.64\scriptsize{$\pm$0.53}} & 
     \textbf{3.01\scriptsize{$\pm$0.12}} & 
     \underline{3.77\scriptsize{$\pm$1.47}} & \underline{1.41\scriptsize{$\pm$0.19}} 
     \\
     \midrule
    
     \multirow{9}{*}{NLL~($\downarrow$)} & \loramle & 3.17\scriptsize{$\pm$0.37} & 2.85\scriptsize{$\pm$0.27} & 1.17\scriptsize{$\pm$0.13} & 0.95\scriptsize{$\pm$0.07} & 0.73\scriptsize{$\pm$0.03} & \underline{0.32\scriptsize{$\pm$0.00}} \\
     
     & \loramap & 2.46\scriptsize{$\pm$0.34} & 2.66\scriptsize{$\pm$0.11} & 0.90\scriptsize{$\pm$0.05} & 1.62\scriptsize{$\pm$0.29} & 0.75\scriptsize{$\pm$0.01} & 0.33\scriptsize{$\pm$0.00} \\
     
     & \loramcd~\cite{gal2016dropout} & 
     2.79\scriptsize{$\pm$0.53} & 2.67\scriptsize{$\pm$0.15} & 1.00\scriptsize{$\pm$0.14} & 1.02\scriptsize{$\pm$0.03} & 0.77\scriptsize{$\pm$0.03} & 0.31\scriptsize{$\pm$0.00} \\
     
     & \loraens~\cite{lakshminarayanan2017simple,balabanov2024uncertainty,wang2023lora} & 2.71\scriptsize{$\pm$0.08} & 2.46\scriptsize{$\pm$0.22} & 0.82\scriptsize{$\pm$0.03} & 1.25\scriptsize{$\pm$0.03} & 1.06\scriptsize{$\pm$0.04} & 0.57\scriptsize{$\pm$0.02} \\

     & \loranaivebbb~\cite{blundell2015BBB} & 
     1.40\scriptsize{$\pm$0.55} & 
     2.23\scriptsize{$\pm$0.04} & 
     0.91\scriptsize{$\pm$0.06} & 
     0.84\scriptsize{$\pm$0.15} & 
    0.66\scriptsize{$\pm$0.05} & 
     \textbf{0.31\scriptsize{$\pm$0.00}}
     \\
     
     & \loralap~\cite{yang2023bayesian} & \textbf{0.60\scriptsize{$\pm$0.00}} & 1.03\scriptsize{$\pm$0.04}\footnotemark[1] & 0.88\scriptsize{$\pm$0.00} & 0.57\scriptsize{$\pm$0.01} & \underline{0.52\scriptsize{$\pm$0.01}} & \textbf{0.31\scriptsize{$\pm$0.00}} \\

    \cmidrule{2-8}

     & \ours~(N=0) & 
     0.91\scriptsize{$\pm$0.10} & 
     1.19\scriptsize{$\pm$0.02} & 
     0.56\scriptsize{$\pm$0.01} &
     0.60\scriptsize{$\pm$0.01} & 
     0.56\scriptsize{$\pm$0.02} & \underline{0.32\scriptsize{$\pm$0.00}} 
     \\

     & \ours~(N=5) & 
     0.68\scriptsize{$\pm$0.01} & 
     \underline{0.90\scriptsize{$\pm$0.01}} & 
     \underline{0.46\scriptsize{$\pm$0.02}} &
     \underline{0.56\scriptsize{$\pm$0.01}} & 
     {0.53\scriptsize{$\pm$0.01}} & \underline{0.32\scriptsize{$\pm$0.00}} 
     \\
     
     & \ours~(N=10) & 
     \underline{0.63\scriptsize{$\pm$0.01}} & 
     \textbf{0.78\scriptsize{$\pm$0.02}} & 
     \textbf{0.40\scriptsize{$\pm$0.01}} & 
     \textbf{0.54\scriptsize{$\pm$0.00}} & 
    \textbf{0.50\scriptsize{$\pm$0.01}} & \textbf{0.31\scriptsize{$\pm$0.00}} 
     \\
    \bottomrule[0.12em]
    \end{tabular}
 }
\end{center}
\label{tab:main-llama}
\vspace{-1em}
\end{table*}

We fine-tune Llama2-7B on six common-sense reasoning tasks: Winogrande-small (WG-S), Winogrande-medium (WG-M)~\cite{wg}, ARC-Challenge (ARC-C)~\cite{arc}, ARC-Easy (ARC-E)~\cite{arc}, OpenBookQA (OBQA)~\cite{obqa}, and BoolQ~\cite{boolq}. For all baseline methods, using the same pre-trained LLM backbone, we maintain consistent hyperparameters across all datasets and do not use additional validation sets to achieve higher performance~(See \appref{app:implementation-dataset} for detailed settings).



\Tabref{tab:main-llama} shows the performance of \ours compared to the baselines, including ACC, ECE, and NLL, on the in-distribution test set with the pre-trained Llama2-7B model. The high ECE and NLL for MLE indicate overconfidence in LLMs during conventional fine-tuning, except for BoolQ due to its large dataset size. 
Simple but popular baselines like MAP, MCD, and ENS show mixed results in terms of NLL and/or ECE, highlighting the challenge of uncertainty estimation during LLM fine-tuning. LAP, the most competitive post-training baseline for uncertainty estimation, significantly reduces NLL and ECE on some datasets but lacks consistent performance, as indicated by its failures on ARC-C and ARC-E.
BBB mitigates the overconfidence issue in LLMs across almost all datasets, showcasing the advantage of jointly optimizing the mean and covariance of the variational weight distributions during fine-tuning. However, there remains considerable room for improvement.


\begin{table*}[t]
\caption{
    \textbf{Performance on in-distribution and out-of-distribution datasets}.
    All the uncertainty estimation methods are applied to the LoRA adapter added upon the pre-trained Llama2-7B weights.
}\vspace{-0.8em}
\begin{center}
\resizebox{1\linewidth}{!}{%
\setlength{\tabcolsep}{10pt}
\begin{tabular}{clccc cc}
	\toprule[0.12em]
	\multirow{3}{*}[-0.5em]{\textbf{Metric}} & \multirow{3}{*}[-0.5em]{\textbf{Method}} & \multicolumn{5}{c}{\textbf{Datasets}} \\
    \cmidrule{3-7}
    & & \emph{In-Dist.} & \multicolumn{2}{c}{\emph{Smaller Dist. Shift}} & \multicolumn{2}{c}{\emph{Larger Dist. Shift}}\\
    \cmidrule(lr){3-3}\cmidrule(lr){4-5}\cmidrule(lr){6-7}
    & & OBQA~\cite{obqa} & ARC-C~\cite{arc} & ARC-E~\cite{arc} & Chem~\cite{mmlu,ethics} & Phy~\cite{mmlu,ethics} \\
    \midrule
    
     \multirow{9}{*}{ACC~($\uparrow$)} & \loramle 
     & 81.52\scriptsize{$\pm$0.25} 
     & 66.20\scriptsize{$\pm$0.87} 
     & 75.12\scriptsize{$\pm$0.85} 
     & 40.62\scriptsize{$\pm$2.25}
     & 28.82\scriptsize{$\pm$1.30} \\
     
     & \loramap 
     & 81.38\scriptsize{$\pm$0.91} 
     & \underline{69.59\scriptsize{$\pm$0.33}}
     & 75.47\scriptsize{$\pm$0.73} 
     & \textbf{44.79\scriptsize{$\pm$0.00}}
     & 28.47\scriptsize{$\pm$1.20} \\
     
     & \loramcd~\cite{gal2016dropout} 
     & 81.72\scriptsize{$\pm$0.10}
     & 69.03\scriptsize{$\pm$0.70} 
     & 76.00\scriptsize{$\pm$1.58} 
     & 42.71\scriptsize{$\pm$0.01}
     & 29.17\scriptsize{$\pm$4.54} \\
     
     & \loraens~\cite{lakshminarayanan2017simple,balabanov2024uncertainty,wang2023lora}
     & 81.38\scriptsize{$\pm$0.65} 
     & 67.34\scriptsize{$\pm$0.70} 
     & 75.18\scriptsize{$\pm$2.03}
     & 43.75\scriptsize{$\pm$1.04}
     & 30.56\scriptsize{$\pm$2.62}
      \\

      & \loranaivebbb~\cite{blundell2015BBB} & 
     82.06\scriptsize{$\pm$0.59} & 
     67.25\scriptsize{$\pm$1.18} & 
     75.83\scriptsize{$\pm$0.75} & 
     42.36\scriptsize{$\pm$0.49} & 
     30.21\scriptsize{$\pm$2.25} \\

     & \loralap~\cite{yang2023bayesian} 
     & \underline{82.12\scriptsize{$\pm$0.67}} 
     & 69.14\scriptsize{$\pm$1.15}
     & 74.94\scriptsize{$\pm$0.96}
     & \underline{44.10\scriptsize{$\pm$1.30}}
     & 31.60\scriptsize{$\pm$0.49} \\

     \cmidrule{2-7}

     & \ours~(N=0)
     & \textbf{82.73\scriptsize{$\pm$0.41}}
     & \textbf{69.93\scriptsize{$\pm$1.20}}
     & \textbf{76.88\scriptsize{$\pm$0.41}}
     & 41.67\scriptsize{$\pm$2.25}
     & \underline{31.94\scriptsize{$\pm$1.77}} \\

     & \ours~(N=5)
     & 81.79\scriptsize{$\pm$0.94}
     & 68.36\scriptsize{$\pm$1.39}
     & 75.82\scriptsize{$\pm$1.15}
     & 40.62\scriptsize{$\pm$3.07}
     & \textbf{32.64\scriptsize{$\pm$0.98}} \\
     
     & \ours~(N=10)
     & 81.52\scriptsize{$\pm$0.74}
     & 67.71\scriptsize{$\pm$1.13}
     & \underline{76.37\scriptsize{$\pm$0.80}}
     & \textbf{44.79\scriptsize{$\pm$1.47}}
     & 31.60\scriptsize{$\pm$2.73} \\
     \midrule

     \multirow{9}{*}{ECE~($\downarrow$)} & \loramle 
     & 12.55\scriptsize{$\pm$0.46} 
     & 22.20\scriptsize{$\pm$0.39} 
     & 16.47\scriptsize{$\pm$0.86} 
     & 21.72\scriptsize{$\pm$0.30}
     & 29.60\scriptsize{$\pm$1.29} \\
     
     & \loramap 
     & 15.34\scriptsize{$\pm$0.27} 
     & 19.31\scriptsize{$\pm$1.46} 
     & 15.68\scriptsize{$\pm$0.51} 
     & 17.55\scriptsize{$\pm$1.95}
     & 30.25\scriptsize{$\pm$2.18} \\
     
     & \loramcd~\cite{gal2016dropout} 
     & 14.45\scriptsize{$\pm$0.84} 
     & 19.54\scriptsize{$\pm$0.33} 
     & 15.32\scriptsize{$\pm$1.16} 
     & 17.9\scriptsize{$\pm$0.63}
     & 29.53\scriptsize{$\pm$4.20} \\
     
     & \loraens~\cite{lakshminarayanan2017simple,balabanov2024uncertainty,wang2023lora}
     & 13.26\scriptsize{$\pm$0.82} 
     & \underline{7.59\scriptsize{$\pm$1.43}} 
     & 6.44\scriptsize{$\pm$0.83} 
     & 12.04\scriptsize{$\pm$4.57}
     & \underline{17.52\scriptsize{$\pm$1.28}}\\

     & \loranaivebbb~\cite{blundell2015BBB} & 
     11.38\scriptsize{$\pm$1.07} & 
     19.90\scriptsize{$\pm$0.66} & 
     13.41\scriptsize{$\pm$0.85} & 
     15.67\scriptsize{$\pm$1.23} & 
     26.10\scriptsize{$\pm$4.76} \\

     & \loralap~\cite{yang2023bayesian} 
     & 8.70\scriptsize{$\pm$1.77} 
     & \textbf{5.84\scriptsize{$\pm$0.64}}
     & 8.51\scriptsize{$\pm$1.06}
     & \underline{10.76\scriptsize{$\pm$3.41}}
     & \textbf{13.91\scriptsize{$\pm$0.90}} \\

     \cmidrule{2-7}

     & \ours~(N=0)
     & 8.36\scriptsize{$\pm$0.38}
     & 14.00\scriptsize{$\pm$1.02}
     & 10.70\scriptsize{$\pm$0.39}
     & 15.05\scriptsize{$\pm$0.77}
     & 22.90\scriptsize{$\pm$2.27} \\

     & \ours~(N=5)
     & \textbf{3.40\scriptsize{$\pm$0.57}}
     & 9.76\scriptsize{$\pm$0.71}
     & \underline{5.96\scriptsize{$\pm$0.93}}
     & 14.33\scriptsize{$\pm$1.55}
     & 18.15\scriptsize{$\pm$1.96} \\
     
     & \ours~(N=10) 
     & \underline{3.77\scriptsize{$\pm$1.47}}
     & 9.55\scriptsize{$\pm$0.40} 
     & \textbf{5.48\scriptsize{$\pm$1.27}}
     & \textbf{9.77\scriptsize{$\pm$1.35}}
     & 18.29\scriptsize{$\pm$1.35} \\
     
     \midrule

     \multirow{9}{*}{NLL~($\downarrow$)} & \loramle 
     & 0.73\scriptsize{$\pm$0.03} 
     & 1.16\scriptsize{$\pm$0.00} 
     & 0.92\scriptsize{$\pm$0.03} 
     & 1.56\scriptsize{$\pm$0.06}
     & 1.66\scriptsize{$\pm$0.05} \\
     
     & \loramap 
     & 1.06\scriptsize{$\pm$0.04} 
     & 1.10\scriptsize{$\pm$0.07} 
     & 0.93\scriptsize{$\pm$0.04} 
     & 1.55\scriptsize{$\pm$0.06}
     & 1.65\scriptsize{$\pm$0.03} \\
     
     & \loramcd~\cite{gal2016dropout} 
     & 1.06\scriptsize{$\pm$0.08} 
     & 1.08\scriptsize{$\pm$0.01} 
     & 0.88\scriptsize{$\pm$0.03} 
     & 1.59\scriptsize{$\pm$0.07}
     & 1.67\scriptsize{$\pm$0.05} \\
     
     & \loraens~\cite{lakshminarayanan2017simple,balabanov2024uncertainty,wang2023lora} 
     & 0.75\scriptsize{$\pm$0.01} 
     & 0.86\scriptsize{$\pm$0.01} 
     & 0.69\scriptsize{$\pm$0.03}
     & \textbf{1.28\scriptsize{$\pm$0.00}}
     & \underline{1.39\scriptsize{$\pm$0.03}}\\

     & \loranaivebbb~\cite{blundell2015BBB} & 
     0.66\scriptsize{$\pm$0.05} & 
     1.06\scriptsize{$\pm$0.01} & 
     0.79\scriptsize{$\pm$0.02} & 
     1.49\scriptsize{$\pm$0.05} & 
     1.62\scriptsize{$\pm$0.06}\\

     & \loralap~\cite{yang2023bayesian} 
     & \underline{0.52\scriptsize{$\pm$0.01}}
     & \textbf{0.81\scriptsize{$\pm$0.00}}
     & 0.70\scriptsize{$\pm$0.02}
     & \underline{1.35\scriptsize{$\pm$0.03}}
     & \textbf{1.36\scriptsize{$\pm$0.01}} \\

     \cmidrule{2-7}

     & \ours~(N=0)
     & 0.56\scriptsize{$\pm$0.02}
     & 0.89\scriptsize{$\pm$0.02}
     & 0.67\scriptsize{$\pm$0.02}
     & 1.44\scriptsize{$\pm$0.00}
     & 1.53\scriptsize{$\pm$0.02} \\

     & \ours~(N=5)
     & 0.53\scriptsize{$\pm$0.01}
     & 0.85\scriptsize{$\pm$0.00}
     & \underline{0.64\scriptsize{$\pm$0.01}}
     & 1.39\scriptsize{$\pm$0.02}
     & 1.48\scriptsize{$\pm$0.01} \\
     
     & \ours~(N=10) 
     & \textbf{0.50\scriptsize{$\pm$0.01}}
     & \underline{0.83\scriptsize{$\pm$0.01}} 
     & \textbf{0.60\scriptsize{$\pm$0.01}}
     & 1.38\scriptsize{$\pm$0.01}
     & 1.46\scriptsize{$\pm$0.02} \\

	\bottomrule[0.12em]
	\end{tabular}
	}
\end{center}
\label{tab:ood-llama}
\vspace{-1.2em}
\end{table*}
\footnotetext[1]{{
LAP encounters training failures on the ARC-C dataset under a unified setting, where the same hyperparameter configuration is applied across all datasets. To achieve competitive performance with LAP on the ARC-C dataset, we deviate from this unified setting, allowing dataset-specific hyperparameters for LAP. Note that this introduces an unfair advantage for LAP. Specifically, for the LAP on ARC-C dataset, we set the dropout rate to 0.1, the learning rate to 5e-5, and applied early stopping at the 5,000-th iteration (out of 10,000).}}
\subsection{Results on In-distribution Datasets}
\label{sec:exp-id}

\ours consistently achieves better or comparable performance across all datasets. With the number of samples during inference set to $N=10$, the same as MCD, \ours provides the best uncertainty estimation performance, significantly reducing NLL and ECE, and greatly mitigating overconfidence while maintaining comparable or better ACC than MLE. Even with half the number of samples, $N=5$, \ours still delivers performance comparable to that of $N=10$ and outperforms other baselines on most datasets. 
{By abandoning the modeling of the posterior distribution, prediction using the mean of the weight distribution, i.e., \ours (N=0) sacrifices some degree of calibration in exchange for improved accuracy. \appref{app:tradeoff} presents the trade-off between accuracy and calibration, which is controlled by the standard deviation of the prior Gaussian distribution).}


Besides Llama2-7B, we also include additional results for RoBERTa-base~\cite{liu2019roberta} on text classification tasks in \appref{app:add-exp-robert}. 
Our method consistently achieved either the best or runner-up performance across nearly all datasets,  demonstrating its versatility across different architectures. 



\subsection{Results on Out-of-Distribution Datasets}
\label{sec:exp-ood}
We use models fine-tuned on OBQA~\cite{obqa} to evaluate the generalization ability of different methods under distributional shifts. OBQA consists of multiple-choice elementary-level science questions. We categorize the distributional shifts into two types: \emph{smaller} and \emph{larger} shifts. The ARC~\cite{arc} dataset, which also consists of multiple-choice science questions, represents a smaller distributional shift. The college-level chemistry and physics subsets of MMLU~\cite{mmlu} represent larger distributional shifts.

The results in \Tabref{tab:ood-llama} highlight \ours's superior OOD generalization ability compared to other methods on both smaller and larger distribution shifts. 
\ours achieves the highest accuracy when solely utilizing the mean of the weight distribution in smaller distribution shifts.
For larger distribution shifts, incorporating uncertainty through sampling improves model accuracy.
Regarding uncertainty estimation, \ours demonstrates the best or second-best performance in smaller distribution shifts. Although there is a slight performance drop with larger distribution shifts, \ours remains comparable to baselines such as ENS and LAP.





\section{Related Work}
\label{sec:related}
\textbf{Parameter-Efficient Fine-Tuning~(PEFT) for LLMs.}
Due to the prohibitively large size of LLMs, parameter-efficient fine-tuning has become a trending topic. Computational paradigms in this area include adapter-based fine-tuning ~\cite{adap1, adap2, adap3, adap4, adap5}, prompt-based fine-tuning~\cite{prompt1, prompt2, prompt3, prefix1, prompt4, prompt5}, and partial fine-tuning~\cite{partial1, partial2, partial3, partial4, partial5}. Among these, LoRA~\cite{hu2022lora} has gained significant attention due to its simplicity and effectiveness. Building on LoRA, numerous studies have aimed to further optimize parameter efficiency when fine-tuning large models~\cite{edalati2022krona,han2023svdiff,ding2023sparse,dettmers2024qlora}. For instance, KronA models weight updates as the Kronecker product of two smaller matrices without decreasing the update rank~\cite{edalati2022krona}, and SVDiff performs Singular Value Decomposition (SVD) on the original weight matrices, fine-tuning only the singular values~\cite{han2023svdiff}. 
However, in this paper, we focus solely on Bayesianizing LoRA due to its widespread application in existing works. We also note that \ours can be naturally adapted to handle different LoRA variants.

\textbf{Uncertainty Estimation in Large Language Models.} 
Large-scale pre-trained models are well-calibrated during pre-training~\cite{kadavath2022language}, but fail to accurately express predictive uncertainty during inference~\cite{amodei2016concrete,weidinger2021ethical,kadavath2022language,huang2023survey,tian2023just}, especially after fine-tuning~\cite{balabanov2024uncertainty,wang2023lora,yang2023bayesian,onal2024gaussian}. This indicates that measures effective during pre-training~\cite{bayesllm1, bayesllm2, bayesllm3, bayesllm4, bayesllm5} may lose their power of uncertainty estimation after fine-tuning for domain-specific knowledge.
To address this issue, \cite{fan2020bayesian,zhang2021bayesian} define priors and approximate posteriors on the full attention weights during fine-tuning, achieving better uncertainty estimation but at a significant cost in time and space. 
Consequently, recent work integrates Bayesian methods and PEFT for efficient uncertainty estimation. For instance, \cite{balabanov2024uncertainty,wang2023lora} train and store multiple copies of different LoRAs, ensembling their outputs during inference to achieve somewhat better results. \cite{yang2023bayesian} applies Kronecker factorized Laplace approximation on fine-tuned LoRA. However, such post-training procedures bifurcate posterior approximation into two stages, leading to suboptimal estimation.
In contrast, our \ours enables simultaneous estimation of both the mean and covariance of LLM parameters in a single fine-tuning stage, substantially improving performance.

\section{Conclusion}
\label{sec:conclusion}

In this work, we propose a principled Bayesianization framework for parameter-efficiently fine-tuning LLMs. 
Our theoretical analysis shows that a full-weight variational distribution can be efficiently optimized by approximately using a low-rank space of the weight update matrices. 
Our empirical evaluations corroborate this theoretical insight, demonstrating superior generalization and uncertainty estimation capabilities across diverse scenarios compared to various baseline methods.
Building on LoRA, our approach seamlessly integrates with existing LLM architectures while imposing minimal additional memory overhead and training time. 
Our method highlights that jointly learning the mean and covariance of the variational distribution during fine-tuning can mutually improve both, underscoring the powerful potential of Bayesian methods in enhancing the reliability and generalization of LLMs.

\section{Limitations}
\label{sec:limitations}
{
The main limitations of our proposed \ours method are:
(i) \ours is confined to fine-tuning scenarios and is not applicable to training-free tasks, such as direct uncertainty estimation during inference~\cite{TFUncertainty}.
(ii) As a typical mean-field variational inference method, \ours requires multiple sampling iterations during inference, which challenges stable and efficient deployment.
(iii) While \ours's effectiveness has been empirically demonstrated for downstream classification tasks, its application to generation tasks requires further investigation.
}

\section*{Acknowledgement}
\label{sec:acknowledgement}
{
The authors thank the reviewers/AC for the constructive comments to improve the paper. 
We thank Sanket Jantre for identifying improvements to our proof and for other valuable discussions.
HS and HW are partially supported by Microsoft Research AI \& Society Fellowship, NSF Grant IIS-2127918, NSF CAREER Award IIS-2340125, NIH Grant 1R01CA297832, and the Amazon Faculty Research Award. 
This research is also supported by NSF National Artificial Intelligence Research Resource (NAIRR) Pilot. 
The views and conclusions contained herein are those of the authors and should not be interpreted as necessarily representing the official policies, either expressed or implied, of the sponsors.
}

{
\bibliographystyle{abbrv}
\bibliography{ref}
}

\appendix
\clearpage
\section*{\LARGE Appendix}
\markboth{Appendix}{Appendix}

In \appref{app:proof}, we present the proofs for the theorems in the main body of our paper. 
In \appref{app:implementation}, we introduce the experimental settings, including evaluation metrics and training schemes. 
Finally, in \appref{app:add-exp}, we present supplementary empirical results including the experiments on another language model and analysis of the space and time cost of our algorithm.

\section{Proof of Theorems and Claims}\label{app:proof}

In this section, we first present the proof of the two main theorems~(\Thmref{thm:posterior} and \Thmref{thm:prior}) in \appref{app:proof-main}.
Next, we show how a analysis on our design of parameterization in \appref{app:reparameter}. 
Finally, we provide a detailed derivation of the LoRA Flipout in \appref{app:flipout}.

\subsection{Proof of Main Theorems}
\label{app:proof-main}

\begin{customThm}{3.1}[\textbf{Variational Distribution of the Full-Weight Matrix in \ours}]
    With the pre-trained weight matrix $\mW_0 \in\mathbb{R}^{m\times n}$ and the low-rank weight update matrix $\mB \in \mathbb{R}^{m\times r}$, suppose that the variational distribution of the other low-rank update matrix $\mA \in \mathbb{R}^{r\times n}$ is Gaussian with $q(\mA|\vtheta=\{\mM, \mOmega\})=\prod_{ij} \gN (A_{ij}|M_{ij},\Omega_{ij}^2)$, where $\mM=[M_{ij}] \in \mathbb{R}^{r\times n}$ and $\mOmega=[\Omega_{ij}] \in \mathbb{R}^{r\times n}$ are its mean and standard deviation, respectively. The equivalent variational distribution defined on the full weight matrix $\mW$ as in \Eqref{eq:vfe} is given by
    \begin{align*}
        q(\vectorize(\mW)|\mB, \vtheta) &= \gN(\vectorize(\mW)|\vmu_q, \mSigma_q), \\
        \text{where }\quad 
        \vmu_q &= \vectorize(\mW_0+\mB\mM), \\
        \mSigma_q &= [\mI_n \otimes \mB] [\diag(\vectorize(\mOmega)^2)] [\mI_n \otimes \mB^\top].  
    \end{align*}
\end{customThm}

\begin{proof}
    We begin by calculating the mean value of $q$,
    \begin{align}
        \vmu_q 
        &= \vectorize(\E[\mW_0+\mB\mA]) \\
        &= \vectorize(\mW_0+\mB\E[\mA]) \\
        &= \vectorize(\mW_0+\mB\mM).
    \end{align}

    Suppose the deterministic matrix $\mB=[\vb_1, \vb_2, \cdots, \vb_r]\in \mathbb{R}^{m\times r}$, random matrix $\mA = [\va_1, \va_2, \cdots, \va_r]^\top \in \mathbb{R}^{r\times n}$, with its underlying parameters of mean and standard deviation defined likewise $\mM\in \mathbb{R}^{r\times n}$ and $\mOmega\in \mathbb{R}^{r\times n}$.
    We have $\mW = \mB\mA = \sum_{i=1}^r \vb_i\va_i^\top$.
    We then rewrite $\vectorize(\mW)$ in the form of Kronecker product $\otimes$:
    \begin{align}
        \vectorize(\mW) 
        &= \vectorize(\sum_{i=1}^r \vb_i\va_i^\top) 
        = \sum_{i=1}^{r}(\va_i \otimes \vb_i)
    \end{align}
    

    We then calculate the covariance matrix $\mSigma_q$ as 
    \begin{align}
        \mSigma_q 
        &= \operatorname{cov}[\vectorize(\mW), \vectorize(\mW)]
        = \operatorname{cov}[\sum_{i=1}^{r}(\va_i \otimes \vb_i), \sum_{i=1}^{r}(\va_i \otimes \vb_i)] \\
        &= \sum_{i=1}^{r}\operatorname{cov}[\va_i \otimes \vb_i, \va_i \otimes \vb_i] +
        \sum_{i\neq j}\operatorname{cov}[\va_i \otimes \vb_i, \va_j \otimes \vb_j] \\
        &= \sum_{i=1}^{r} \left\{\E_{\va_i}[(\va_i \otimes \vb_i)(\va_i \otimes \vb_i)^\top] - \E_{\va_i}[(\va_i \otimes \vb_i)]\E_{\va_i}[(\va_i \otimes \vb_i)^\top] \right\}\\
        &= \sum_{i=1}^{r} \left\{\E_{\va_i}[(\va_i \va_i^\top)] \otimes (\vb_i \vb_i^\top) - (\E_{\va_i}[\va_i]\E_{\va_i}[\va_i]^\top) \otimes (\vb_i\vb_i^\top)\right\} \\
        &= \sum_{i=1}^{r} \operatorname{diag}(\vsigma_i^2) \otimes (\vb_i \vb_i^\top) \\
        &= [\mI_n \otimes \mB] [\diag(\vectorize(\mOmega^2))] [\mI_n \otimes \mB^\top],
    \end{align} 
completing the proof. 
\end{proof}

It is crucial to note here, the final covariance matrix of $q(\vectorize(\mW))$ follows a block-diagonal structure, which will be further utilized for the proof of \Thmref{thm:prior}. Defining $\mSigma_i = \operatorname{diag}(\mOmega_{i:}^2)$, we have:
\begin{align}
    \mSigma_q &=
    \begin{bmatrix}
        \mB\mSigma_1\mB^\top & \\
        & \ddots & \\
        & & \mB\mSigma_n\mB^\top
    \end{bmatrix}.
\end{align}

Another important fact about $\mSigma_q$ is its singularity. It can be seen directly as we consider the rank of any one of the block matrix $\mB\mSigma_i\mB^\top \in \mathbb{R}^{m \times m}, \forall i\in [n]$: 
\begin{align}
    \operatorname{r}(\mB\mSigma\mB^\top) &\leq \min\{\operatorname{r}(\mB),\operatorname{r}(\mSigma_i),\operatorname{r}(\mB^\top)\} \leq r < m, 
\end{align}
where $r$ is the rank of LoRA, strictly smaller than the output dimension of $m$.

\begin{customThm}{3.2}[\textbf{Efficient Computation of Full-Weight KL Divergence}]
    Suppose the pre-trained weights $\mW_0$, update matrix $\mB$, and the variational distribution $q(\mA|\vtheta)$ are defined as in \Thmref{thm:posterior}, 
    and the prior distribution of the full-weight matrix $P(\vectorize(\mW))$ is defined as \Eqref{eq:prior}. 
    Consider the Gaussian prior distribution 
    $P(\mA)=\prod_{ij} \gN (A_{ij}|0,\sigma_p^2)$; we then have:
    \begin{align*}
        \operatorname{KL}[ q(\vectorize(\mW)|\mB,\vtheta) \| P(\vectorize(\mW))]
        &= \operatorname{KL}[ q(\mA|\vtheta) \| P(\mA)],
    \end{align*}
    if $\tilde{\mR}=[\sigma_p \mI_n \otimes \mR]$, where $\mR$ satisfies $\mR\mR^\top = \mB\mB^\top$.
\end{customThm}

\begin{proof}
    We start by assuming the low-rank structure of the prior $P(\vectorize(\mW))$, and then reveal the conditions reaching to our final conclusion step by step. 

    Typically, for two Gaussian distributions $q$ and $p$ whose covariance matrices $\mSigma_q\in \mathbb{R}^{d\times d}$ and $\mSigma_p\in \mathbb{R}^{d\times d}$ are both full-rank, and their means as $\vmu_q\in \mathbb{R}^{d}$ and $\vmu_p\in \mathbb{R}^{d}$, we have their KL-divergence as
    \begin{align}\label{eq:kl}
        \operatorname{KL}[q \| p] &= \frac{1}{2} \left[ 
            \log \frac{|\mSigma_p|}{|\mSigma_q|} 
            - d 
            + \operatorname{tr}(\mSigma_p^{-1}\mSigma_q) 
            + (\vmu_q - \vmu_p)^\top \mSigma_p^{-1} (\vmu_q - \vmu_p)
        \right].
    \end{align}

    The singularity of the covariance matrices of $P(\vectorize(\mW))$ and $q(\vectorize(\mW))$, i.e., $|\mSigma_q|=|\mSigma_p|=0$, can cause issues when computing the KL-divergence as it includes the log-determinant term. Therefore in this proof, we consider the alternative of the covariance matrices, where an extremely small diagonal elements are added. 
    
    \textbf{For the prior distribution}, following the alternative form of a degenerate Gaussian~\cite{schoeman2022degenerate}, as suggested in \eqnref{eq:prior}, we assume  
    \begin{align}
        P(\vectorize(\mW)) &= \gN (\mW_0, \mSigma_p), \\
        \nonumber\text{where }\qquad \mSigma_p &= \lambda \mI + \tilde{\mR}\tilde{\mR}^\top, \quad(\lambda \rightarrow 0^+).
    \end{align}
    By default, we assume that the low-rank tall matrix $\tilde{\mR}\in \mathbb{R}^{(mn)\times r^\prime}$ has the full column rank $r^\prime$. Otherwise if $\operatorname{r}(\tilde{\mR}) = r^{\prime\prime} < r^\prime$, then we can in effect consider a new matrix component $\tilde{\mR}^\prime\in \mathbb{R}^{(mn)\times r^{\prime\prime}}$ that has the same rank as $r^{\prime\prime}$, which satisfies our assumption of full column rank. 
    Therefore, we have the SVD decomposition of $\tilde{\mR}$ is given by
    \begin{align}\label{eq:svd}
        \tilde{\mR} &= \mU_R \mD_R \mV_R^\top,
    \end{align}
    where $\mU_R\in\mathbb{R}^{(mn)\times (mn)}$ and $\mV_R\in\mathbb{R}^{r^\prime\times r^\prime}$ are orthonormal, i.e., $\mU_R\mU_R^\top = \mU_R^\top\mU_R = \mI_{(mn)}$ and $\mV_R\mV_R^\top = \mV_R^\top\mV_R = \mI_{r^\prime}$. 
    $\mD_R$ is a tall matrix where its upper part is diagonal and the lower part is a zero matrix, denoted as $\mD_R = [\mD_R^*, \mO]^\top = [\operatorname{diag}([d_{R_1}>0, d_{R_2}>0, \cdots, d_{R_{r^\prime}}>0]), \mO]^\top$.

    \textbf{For the approximate posterior} $q(\vectorize(\mW)|\mB, \vtheta)$, we consider
    \begin{align}
        q(\vectorize(\mW)|\mB, \vtheta) &= \gN (\mW_0 + \mB\mM, \mSigma_q), \\
        \nonumber\text{where }\qquad \mSigma_q &= \lambda \mI + \tilde{\mB}\mSigma\tilde{\mB}^\top, \quad (\lambda \rightarrow 0^+), 
    \end{align}
    where we simplify the notation for the covariance matrix $\mSigma_q$ by defining $\tilde{\mB} = [\mI_n \otimes \mB]\in\mathbb{R}^{(mn)\times(mr)}$ and $\mSigma = \operatorname{diag}(\vectorize(\mOmega)^2)$. 
    Likewise, we have the SVD-decomposed matrices for $\tilde{\mB}$ where they are defined in the same way as \eqnref{eq:svd}:
    \begin{align}
        \tilde{\mB} &= \mU_B \mD_B \mV_B^\top,
    \end{align}
    where $\mU_B$ and $\mV_B$ are orthogonal matrices, and $\mD_B=[\mD_B^*, \mO]^\top = [\operatorname{diag}([d_{B_1}>0, d_{B_2}>0, \cdots, d_{B_{mr}}>0]), \mO]^\top$.

    First, we calculate the log-determinant part of the KL-divergence. For the log-determinant of the covariance matrix of the prior distribution $\mSigma_p$, by applying SVD decomposition in \eqnref{eq:svd}, we have 
    \begin{align}
        \log |\mSigma_p| 
        &= \log |\lambda\mI + \tilde{\mR}\tilde{\mR}^\top|
        = \log | \lambda\mI + \mU_R\mD_R\mV_R^\top \mV_R\mD_R^\top\mU_R^\top | \\
        &= \log |\mU_R (\lambda \mI + \mD_R\mD_R^\top) \mU_R^\top| \\
        &= \log \left| 
        \mU_R 
        \begin{bmatrix}
            (\mD_R^*)^2 + \lambda \mI_{r^\prime} & \mO \\
            \mO & \lambda \mI_{mn-r^\prime}
        \end{bmatrix} 
        \mU_R^\top 
        \right| \\
        &= \log | (\mD_R^*)^2+\lambda \mI_{r^\prime} | + \log |\lambda \mI_{mn-r^\prime}| \\ 
        &= (mn-r^\prime)\log\lambda + \sum_{i=1}^{r^\prime}\log(d_{R_i}^2+\lambda).
    \end{align}
    Following (almost) the same idea, we now have the log-determinant of the variational distribution's covariance $\mSigma_q$ as 
    \begin{align}
        \log |\mSigma_q| 
        &= \log |\lambda \mI + \tilde{\mB}\tilde{\mB}^\top|
        = \log 
        \left| 
        \begin{matrix}
            \mD_B^*\mV_B^\top \mSigma \mV_B\mD_B^* + \lambda \mI_{mr} & \mO \\
            \mO & \lambda \mI_{mn-mr}
        \end{matrix} 
        \right| \\
        &= (mn-mr)\log \lambda + 2\log |\mD_B^*| + \log |\mV_B^\top \mSigma \mV_B + \lambda (\mD_B^*)^{-2} | \\
        &= (mn-mr)\log \lambda + 2\sum_{i=1}^{mr}\log d_{B_i} + \log |\mSigma| + \log |\mI + \lambda \mV_B^\top \mSigma^{-1} \mV_B (\mD_B^*)^{-2} |. \label{eq:log-cov-posterior}
    \end{align}
    We make two observations when $\lambda \rightarrow 0^+$: (i)~compare the terms that contain $\log\lambda$ on both sides, to make sure the log-determinant in the divergence term \emph{bounded}, we have to set $r^\prime = mr$; (ii)~the last term in \eqnref{eq:log-cov-posterior}, $\log |\mI + \lambda \mV_B^\top \mSigma^{-1} \mV_B (\mD_B^*)^{-2} | = \log|\mI|=0$. 
    Therefore, we have
    \begin{align}
        \label{eq:log-det}
        \log \frac{|\mSigma_p|}{|\mSigma_q|}
        &= 
        \sum_{i=1}^{mr} \log\frac{d_{R_i}^2+\lambda}{d_{B_i}^2} 
        - \log|\mSigma|.
    \end{align}

    Next we calculate $\operatorname{tr}(\mSigma_p^{-1}\mSigma_q)$ in \eqnref{eq:kl}. 
    Following the same assumptions and notations above, we have the inverse of the covariance of the prior distribution as
    \begin{align}
        \mSigma_p^{-1} &= \mU_R 
        \begin{bmatrix}
            [(\mD_R^*)^2 + \lambda \mI]^{-1} & \mO \\
            \mO & \lambda^{-1} \mI
        \end{bmatrix} \mU_R^\top.
    \end{align}
    Hence we have
    \begin{align}\label{eq:trace-1}
        \operatorname{tr}(\mSigma_p^{-1}\mSigma_q) 
        &= \operatorname{tr} (
            \mU_R \begin{bmatrix}
            [(\mD_R^*)^2 + \lambda \mI]^{-1} & \mO \\
            \mO & \lambda^{-1} \mI
        \end{bmatrix} \mU_R^\top 
        \mU_B 
        \begin{bmatrix}
            \mD_B^*\mV_B^\top\mSigma\mV_B\mD_B^* + \lambda \mI & \mO \\
            \mO & \lambda\mI
        \end{bmatrix} \mU_B^\top
        ).
    \end{align}

    By using the condition $\mR\mR^\top=\mB\mB^\top$ and $\tilde{\mR}=[\sigma_p\mI_n\otimes\mR]$ defined in \Thmref{thm:prior}, we have
    \begin{align}
        \tilde{\mR}\tilde{\mR}^\top = \sigma_p^2 \tilde{\mB}\tilde{\mB}^\top,
    \end{align}
    and there exists an orthogonal matrix $\mP\in \mathbb{R}^{(mr)\times (mr)}$, such that
    \begin{align}
        \tilde{\mR} &= \sigma_p \tilde{\mB}\mP.
    \end{align}
    The SVD decomposition on $\tilde{\mR}$ can then be formulated as: 
    \begin{align}
        \tilde{\mR} 
        &= \sigma_p \mU_B \mD_B \mV_B^\top \mP \\
        &= (\mU_R=\mU_B) (\mD_R=\sigma_p\mD_B) (\mV_R=\mV_B^\top \mP).
    \end{align}
    Substituting $\mU_R, \mD_R, \mV_R$ back to \eqnref{eq:trace-1} and applying $\lambda\rightarrow 0^+$, we have 
    \begin{align}
        \operatorname{tr}(\mSigma_p^{-1}\mSigma_q) 
        &= \operatorname{tr}(\mI_{mn-nr}) + \operatorname{tr}([(\sigma_p\mD^*)^2+\lambda\mI]^{-1}[\mD_B^*\mV_B^\top\mSigma\mV_B\mD_B^*]) \\
        &= (mn-nr) + \sigma_p^{-2}\operatorname{tr}(\mV_B^\top\mSigma\mV_B) \\ 
        &= (mn-nr) + \sigma_p^{-2}\operatorname{tr}(\mSigma).\label{eq:trace}
    \end{align}

    For the quadratic term in \eqnref{eq:kl}, the pre-trained weights $\mW_0$ cancel out, and we can calculate it as 
    \begin{align}
        &\vectorize(\mB\mM)^\top\mSigma_p^{-1}\vectorize(\mB\mM) \\
        =& [\mM_{:1}^\top\mB^\top, \cdots, \mM_{:1}^\top\mB^\top] 
        \begin{bmatrix}
            (\mB\mSigma_1\mB^\top)^{-1} & & \\
            & \ddots & \\
            & & (\mB\mSigma_n\mB^\top)^{-1}
        \end{bmatrix}
        \begin{bmatrix}
            \mB\mM_{:1} \\
            \vdots \\
            \mB\mM_{:n} 
        \end{bmatrix} \\
        =& \sum_{i=1}^n \mM_{:i}^\top\mB^\top (\mB\mSigma_i\mB^\top)^{-1} \mB\mM_{:i} \\
        =& \sum_{i=1}^n \mM_{:i}^\top ( \mV_B 
        \begin{bmatrix}
            \frac{d_{B_1}^2}{\sigma_p^2(d_{B_1}^2+\lambda)} & & \\
            & \ddots & \\
            & & \frac{d_{B_{mr}}^2}{\sigma_p^2(d_{B_{mr}}^2+\lambda)}
        \end{bmatrix} \mV_B^\top
        ) \mM_{:i}  \\
        =& \frac{1}{\sigma_p^2} \sum_{i=1}^n \mM_{:i}^\top \mM_{:i} \\
        =& \frac{1}{\sigma_p^2} \|\mM\|_2^2. \label{eq:quadratic}
    \end{align}

    Finally, proof is completed by combining \eqnref{eq:log-det}, \eqnref{eq:trace}, and \eqnref{eq:quadratic}.
\end{proof}

\subsection{Analysis on \ours Parameterization}
\label{app:reparameter}

\textbf{General Analysis on Parameterization.} 
Consider a path of parameterization for a single variable:
\begin{align}
    \rho\rightarrow\sigma = f(\rho)\rightarrow \gL = l(\sigma),
\end{align}
where $\rho$ is the real parameter we perform update on, $f$ is our parameterization choice for the variable $\sigma$, and $l$ represents the loss function we aim to minimize.
When comparing two different parameterization methods, we consider the same initial conditions of $\sigma = \sigma_0$, and we assume the same step size $\eta$ on the real parameter $\rho$.
To show the influence of the choice of parameterization, we calculate the decrease of the loss value by performing one step of gradient descent. 
First, by the chain rule, the gradient w.r.t. $\rho_0$ is calculated as 
\begin{align}
    \frac{\rm{d} \gL}{\rm{d}\rho}\big|_{\rho_0} 
    &= \frac{\rm{d} \gL}{\rm{d}\sigma} \big|_{\sigma_0} \cdot 
    \frac{\rm{d} \sigma}{\rm{d}\rho} \big|_{\rho_0} 
    = l^\prime(\sigma_0) \cdot f^\prime(\rho_0).
\end{align}
After one step of the gradient descent, we have $\rho_1$ as 
\begin{align}
    \rho_1 &= \rho_0 - \eta \cdot l^\prime(\sigma_0) \cdot f^\prime(\rho_0),
\end{align}
and the loss value decreased at $\rho_1$ can be approximated by the first-order Taylor expansion, 
\begin{align}
    \Delta \gL &= l(f(\rho_0)) - l(f(\rho_1)) \\
    &= l(f(\rho_0)) - [l(f(\rho_0 - \eta \cdot l^\prime(\sigma_0) \cdot f^\prime(\rho_0)))] \\
    &\approx l(f(\rho_0)) - [l(f(\rho_0) - \eta \cdot l^\prime(\sigma_0) \cdot (f^\prime(\rho_0))^2 )] \\
    &\approx \eta \cdot (l^\prime(\sigma_0))^2 \cdot (f^\prime(\rho_0))^2.
\end{align}
Since the initialization of the different parameterization variable $\rho_0$ is set to ensure the same initial condition of $\sigma_0$ for different $f$s, we can see that the amount the loss decreases by after one step of update is proportional to the squared gradient $\Delta \gL \propto(l^\prime(\sigma_0))^2 \cdot (f^\prime(\rho))^2 = (\nicefrac{\rm{d} \gL}{\rm{d} \rho})^2$.

\textbf{Parameterization with $\log(1+\exp(\cdot))$ or $(\cdot)^2$?} 
Previous VBNs~\cite{blundell2015BBB, krishnan2020specifying} typically use a softplus function $\sigma_q = \log(1+\exp(\rho))$ to parameterize the standard deviation. For a single element $\rho$, the derivative of the closed-form solution of the KL divergence in \eqnref{eq:closed costf} is calculated as
\begin{align}
\label{eq:partial1}
\frac{\rm{d} \operatorname{KL}}{\rm{d} \rho} = -\frac{e^\rho}{(1+e^\rho)\log(1+e^\rho)} + \frac{e^\rho\log(1+e^\rho)}{\sigma_p^2(1+e^\rho)} .
\end{align}
Due to the fact that $\sigma_q$ is typically initialized to a small value close to 0 to ensure stable optimization of the likelihood cost term (e.g., $1e-3$), and in order to ensure that the model obtains good uncertainty, $\sigma_p$ is usually set to a larger value close to 1 (e.g., $1e-1$). In this case, the derivative of $\rho$ in \eqnref{eq:partial1} is almost always a constant $-1$, which, based on our previous analysis, leads to slow convergence for large $\sigma_p$ values when $\sigma_q$ is small. 

Therefore, we parameterize $\sigma_q$ with quadratic function: $\sigma_q = \rho^2$. In this case, the derivative of the KL divergence with respect to $\rho$ in \eqnref{eq:closed costf} becomes:
\begin{align}
\label{eq:partial2}
\frac{\rm{d} \operatorname{KL}}{\rm{d} \rho} = -\frac{2}{\rho} + \frac{2\rho^3}{\sigma_p^2}. 
\end{align}
Under the same initialization conditions, the derivative in \eqnref{eq:partial2} is approximately of the order of $\rho^{-1}$, leading to rapid convergence towards larger $\sigma_p$ values when $\sigma_q$ is small.
Building on this, we use SGD without momentum to optimize the complexity loss term, thereby achieving the natural convergence of  $\sigma_q$.

\begin{figure}[h]
    \centering
        \includegraphics[width=0.5\linewidth]{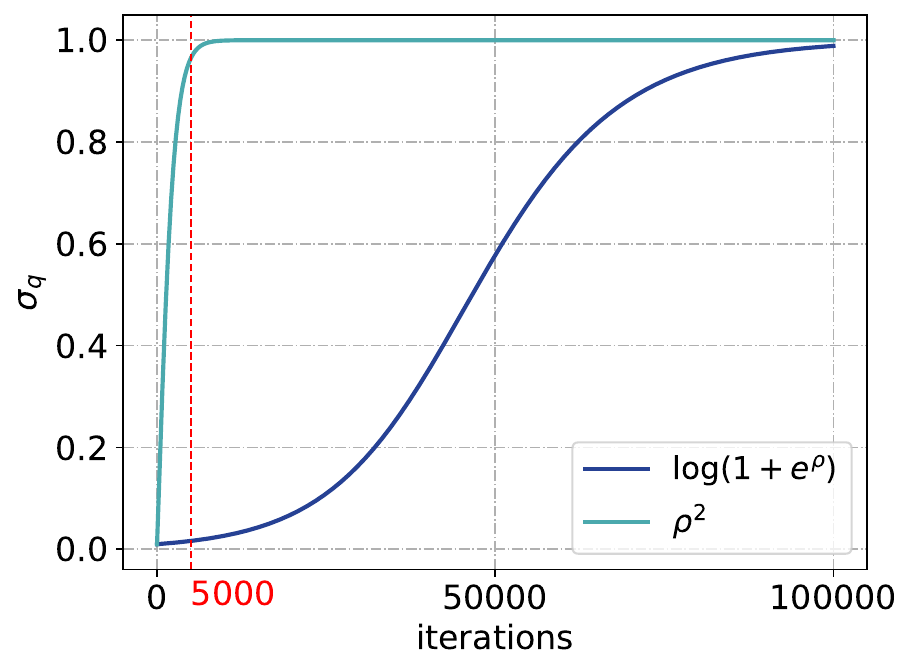} 
        \caption{The growth curve of $\sigma_q = \text{log}(1+e^{\rho})$ and $\sigma_q = \rho^2$ during the optimization of KL divergence~(without data likelihood). The number of gradient steps~(5000) is marked with the red line.}
        \label{fig:repra-softplus}
    \label{fig:repra}
\end{figure}

\textbf{Visualization.}
To visually demonstrate the differences in the convergence of KL divergence during training with these two parameterizations, we set $\sigma_p = 1$ and employ gradient descent to optimize the KL divergence. As introduced in \ref{kl-reweighting}, in practical mini-batch gradient descent, the KL divergence is weighted by {$\nicefrac{1}{\text{\#mini-batches}}$}.  Therefore, assuming there are 100 mini-batches, the learning rate is set to 0.01, which translates to an actual learning rate of $1e-4$ for $\rho$. We initialize $\sigma_q = 0.01$ for both parameterizations. The growth of $\sigma_q$ during the KL training, for $\sigma_q = \text{log}(1+e^{\rho})$ and $\sigma_q = \rho^2$ is shown in~\figref{fig:repra}. In the same setting, the softplus parameterization takes nearly 100,000 gradient steps to converge, while the square parameterization takes only about 5,000 gradient steps. {This modification makes it suitable for our fine-tuning setup, where the number of gradient steps is relatively small.}

\subsection{Deriving Flipout for \ours}
\label{app:flipout}
For the $i$-th input vector $\vh_i$ in a mini-batch, we randomly sample two flipping vector $\vs \in \{-1, +1\}^n$ and $\vt \in \{-1, +1\}^r$. Denoting $\mA$ as the weight matrix sampled from posterior distribution, and $\Delta\mA$ as the batched noise for sampling $\mA$, the output vector $\vz_i$ after applying flipout is:
\begin{align}
    \vz_i 
    &= \mW\vh_i \\
    &= \mW_0\vh_i + \mB\mA\vh_i \\
    &= \mW_0\vh_i + \mB(\mM+\Delta \mA)\vh_i \\
    &= \mW_0\vh_i + \mB\mM\vh + \mB(\hat{\mA}\circ \vt_i\vs_i^\top)\vh_i.
\end{align}
Similarly, for a mini-batch input matrix $\mH \in \mathbb{R}^{n \times b}$ with batch size $b$, we randomly sample two low-rank flipping matrices $\mS\in \{-1,+1\}^{n\times b}$ and $\mT\in \{-1,+1\}^{b\times r}$. The batched output matrix $\mZ$ after applying flipout is then:
\begin{align}
    \mZ &= \mW_0 \mH + \mB (\mM \mH + \left[ \hat{\mA}(\mH \circ \mS) \right] \circ \mT ) \\
    &= \mW_0 \mH + \mB (\mM \mH + \left[ (\mE \circ \mOmega) (\mH \circ \mS) \right] \circ \mT ).
\end{align}

\section{Implementation Details}
\label{app:implementation}

In this section, we first introduce the implementation details of \ours in \appref{app:implementation-ours}, including the KL Re-weighting scheme, initialization of the parameters, and learning scheduling, etc.
Next, we introduce the two evaluation metrics for uncertainty estimation in \appref{app:implementation-metrics}. 
Finally, we present some statistics of the adopted datasets in \appref{app:implementation-dataset}.

\subsection{Implementation of \ours}
\label{app:implementation-ours}

\textbf{KL Re-weighting.}
\label{kl-reweighting}
In mini-batch SGD, the training data $\gD$ is randomly divided into $M$ equally sized subsets: $\gD_1, \gD_2, \ldots, \gD_M$. For mini-batch $i = 1, 2,\ldots,M$, the cost function is:
\begin{align}
    \gF(\gD_i, \vtheta) = - \mathbb{E}_{q(\mW|\vtheta)}[\log P(\gD_i|\mW)] + \lambda_i \operatorname{KL}[q(\mW|\vtheta) \parallel P(\mW)],
\end{align}
where $\lambda_i \in [0, 1]$ and $\sum_{i=1}^{M} \lambda_i = 1$. 
There are various approaches for controlling the weight of KL divergence. \cite{graves2011practical} utilizes $\lambda_i = \nicefrac{1}{M}$, while \cite{blundell2015BBB} adopts $\lambda_i = \nicefrac{2^{M-i}}{2^M-1}$. In fine-tuning tasks, we found that using a scheduler with $\lambda_i = \nicefrac{2^i}{2^M-1}$ performs well. This allows the model to find good fits to the data points within the early stages and then optimize the complexity cost in later stages.

In multiple epochs of mini-batch SGD, larger datasets require more iterations to complete one epoch, resulting in delayed convergence of the complexity cost.
To enhance the stability of \ours's performance across datasets with varying sizes, we pseudo-rescaled the size of the training dataset to make smaller datasets slightly larger and larger datasets slightly smaller. 
For the portions of the dataset that required expansion, we incorporated additional mini-batches from subsequent epochs. Conversely, for the datasets needing reduction, we deferred the excess mini-batches to subsequent epochs.
We denote the size of original dataset as $L_0$, The rescaled dataset size is:
\begin{align}
L^\ast = 100 \cdot L_0 ^ {\frac{\pi}{\gamma}}, 
\end{align}
where $\gamma$ is a coefficient used to control the scaling magnitude, and we set it to 8 in all experiments.

Pseudo-rescaling does not affect the likelihood cost in practical mini-batch gradient descent. It modifies the warm-up period in KL reweighting from $M$ to $\nicefrac{L^\ast}{\text{batch size}}$, enabling more consistent optimization of the complexity cost across datasets of different sizes.
Pseudo-rescaling is specifically designed for training procedures that use a fixed number of iterations across different datasets. However, for training procedures with a fixed number of epochs across different datasets, pseudo-rescaling can be omitted since the complexity cost converges at an identical rate across all datasets.

In \Tabref{tab:5epochs-llama}, we present the results of \ours without pseudo-rescaling after 5 epochs of LoRA fine-tuning, alongside other baseline methods. In this setting, we increase the maximum learning rate from 1e-4 to 2e-4. Fine-tuning for 5 epochs reduces the number of gradient steps required for methods like MLE on smaller datasets such as WG-S, resulting in less overfitting (as evidenced by lower ECE and NLL) while maintaining comparable accuracy to fine-tuning with 5000 gradient steps. \ours continues to demonstrate superior uncertainty estimation performance across nearly all datasets.

\begin{table}[t]
    \begin{minipage}[t]{0.45\linewidth}
        \centering
        \resizebox{1\linewidth}{!}{
        \begin{tabular}{c c c}
        \toprule[0.12em]
        \multirow{2}{*}[-0.25em]{\textbf{Hyperparameter}} 
        & \multicolumn{2}{c}{\textbf{Model}} \\
        \cmidrule{2-3}
        & \textbf{Roberta-base} 
        & \textbf{Llama2-7B} \\
        \midrule
            Optimizer & 
            \multicolumn{2}{c}{AdamW}\\
        \midrule
            LR Scheduler & 
            \multicolumn{2}{c}{Linear}\\
        \midrule
            Warmup Ratio & 
            \multicolumn{2}{c}{0.06}\\
        \midrule
            Learning Rate & 
            $5e-4$ & 
            $1e-4$ \\
        \midrule
            Batch Size & 
            32 & 
            4 \\
        \midrule
            Max Seq. Len. & 
            512 &
            300 \\
        \midrule
            LoRA $\alpha$ & 
            8 &
            16 \\
        \midrule
            LoRA $r$ & 
             \multicolumn{2}{c}{8}\\
        \bottomrule
    \end{tabular}
        }
        \vspace{1em}
        \caption{Hyperparameters of LoRA}
        \label{table.hyperp-base}
    \end{minipage}
\hspace{0.05\linewidth} 
    \begin{minipage}[t]{0.45\linewidth}
        \centering
        \resizebox{1\linewidth}{!}{
        \begin{tabular}{c c c}
        \toprule[0.12em]
        \multirow{2}{*}[-0.25em]{\textbf{Hyperparameter}} 
        & \multicolumn{2}{c}{\textbf{Model}} \\
        \cmidrule{2-3}
        & \textbf{Roberta-base} 
        & \textbf{Llama2-7B} \\
        \midrule
            Optimizer of $\operatorname{KL}$ & 
            \multicolumn{2}{c}{SGD}\\
        \midrule
            LR of $\operatorname{KL}$ & 
            0.002 & 
            0.01\\
        \midrule
            $\sigma_p$ & 
            \multicolumn{2}{c}{0.2}\\
        \midrule
            $\epsilon$ & 
            \multicolumn{2}{c}{0.05}\\
        \midrule
           $\gamma$ & 
            \multicolumn{2}{c}{8}\\
        \bottomrule
        \end{tabular}
        }
        \vspace{1em}
        \caption{\ours-Specific Hyperparameters}
        \label{table.hyperp-blob}
    \end{minipage}%

\end{table}
\begin{table*}[h]
\caption{
    \textbf{Comparison of BLoB without Pseudo-rescaling against baseline
methods applied to LoRA on Llama2-7B pre-trained weights,} {where Accuracy~(\textbf{ACC}) and Expected Calibration Error~(\textbf{ECE}) are reported in percentages.} The evaluation is done across six common-sense reasoning tasks with a shared hyper-parameter setting after \textbf{5 epochs}. 
    We use $N$ to represent the number of samples during inference in \ours. 
    ``$\uparrow$'' and ``$\downarrow$'' indicate that higher and lower values are preferred, respectively. 
    \textbf{Boldface} and \underline{underlining} denote the best and the second-best performance, respectively. 
}
\begin{center}
\resizebox{1\linewidth}{!}{%
\setlength{\tabcolsep}{8pt}

\begin{tabular}{clccc ccc}
	\toprule[0.12em]
	\multirow{2}{*}[-0.25em]{\textbf{Metric}} & \multirow{2}{*}[-0.25em]{\textbf{Method}} & \multicolumn{6}{c}{\textbf{Datasets}}
     \\
     \cmidrule{3-8}
     & & WG-S~\cite{wg}
     & ARC-C~\cite{arc} 
     & ARC-E~\cite{arc} 
     & WG-M~\cite{wg} 
     & OBQA~\cite{obqa} 
     & BoolQ~\cite{boolq} \\
     \midrule

     \multirow{7}{*}{ACC~($\uparrow$)} & \loramle & 68.56\scriptsize{$\pm$0.52} & \textbf{71.39\scriptsize{$\pm$0.52}} & 86.49\scriptsize{$\pm$0.42} & \underline{76.81\scriptsize{$\pm$0.17}} & 81.32\scriptsize{$\pm$0.80} & 88.09\scriptsize{$\pm$0.21}  \\
     
     & \loramap & 68.96\scriptsize{$\pm$0.65} & 69.37\scriptsize{$\pm$0.57} & 86.27\scriptsize{$\pm$0.25} & \textbf{77.35\scriptsize{$\pm$1.11}} & 82.33\scriptsize{$\pm$0.41} & 87.95\scriptsize{$\pm$0.08}  \\
     
     & \loramcd~\cite{gal2016dropout} & \textbf{69.15\scriptsize{$\pm$1.23}}  & 68.81\scriptsize{$\pm$0.16}  & \underline{87.50\scriptsize{$\pm$0.63}}  & 76.64\scriptsize{$\pm$0.89}  & 81.65\scriptsize{$\pm$0.75}  & 88.04\scriptsize{$\pm$0.59} \\
     
     & \loraens~\cite{lakshminarayanan2017simple,balabanov2024uncertainty,wang2023lora} & 67.88\scriptsize{$\pm$0.79}  & 69.14\scriptsize{$\pm$1.11}  & \textbf{87.73\scriptsize{$\pm$0.17}}  & 76.19\scriptsize{$\pm$1.11}  & 82.06\scriptsize{$\pm$0.57}  & \underline{88.10\scriptsize{$\pm$0.17}}  \\

     
     & \loralap~\cite{yang2023bayesian} & \underline{69.01\scriptsize{$\pm$0.49}} & 58.22\scriptsize{$\pm$6.85} & 66.49\scriptsize{$\pm$9.56} & 76.53\scriptsize{$\pm$1.64} & 82.66\scriptsize{$\pm$0.49} & 87.96\scriptsize{$\pm$0.15} \\

    \cmidrule{2-8}
    
     & \ours~(N=0)  & 68.08\scriptsize{$\pm$1.18} & \underline{70.42\scriptsize{$\pm$2.23}} & 86.85\scriptsize{$\pm$0.30} & 74.89\scriptsize{$\pm$1.19} & \textbf{82.99\scriptsize{$\pm$0.69}} & \textbf{88.60\scriptsize{$\pm$0.35}}
     \\
     
     & \ours~(N=10) & 66.69\scriptsize{$\pm$1.12} & 70.00\scriptsize{$\pm$1.80} & 86.37\scriptsize{$\pm$0.47} & 75.17\scriptsize{$\pm$0.76} & \underline{82.92\scriptsize{$\pm$0.74}} &  88.03\scriptsize{$\pm$0.27}  
     \\
     \midrule
    
     \multirow{7}{*}{ECE~($\downarrow$)} & \loramle & 19.56\scriptsize{$\pm$0.56} & 22.49\scriptsize{$\pm$1.11} & 11.33\scriptsize{$\pm$0.51} & 18.37\scriptsize{$\pm$0.52} & 15.60\scriptsize{$\pm$0.32} & 9.65\scriptsize{$\pm$0.07}  \\
     
     & \loramap  &  19.03\scriptsize{$\pm$1.86} & 24.15\scriptsize{$\pm$0.90} & 11.50\scriptsize{$\pm$0.35} & 17.29\scriptsize{$\pm$0.64} & 14.33\scriptsize{$\pm$0.78} & 9.88\scriptsize{$\pm$0.13} \\
     
     & \loramcd~\cite{gal2016dropout}  & 18.69\scriptsize{$\pm$1.37}  & 23.94\scriptsize{$\pm$1.02}  & 10.34\scriptsize{$\pm$0.36}  & 16.78\scriptsize{$\pm$0.41}  & 14.51\scriptsize{$\pm$0.99}  & 9.25\scriptsize{$\pm$0.58} \\
     
     & \loraens~\cite{lakshminarayanan2017simple,balabanov2024uncertainty,wang2023lora} & 14.86\scriptsize{$\pm$3.11}  & 21.86\scriptsize{$\pm$0.94}  & \underline{9.36\scriptsize{$\pm$0.36}}  & 9.21\scriptsize{$\pm$2.16}  & \underline{12.30\scriptsize{$\pm$0.90}}  & 7.65\scriptsize{$\pm$0.02} \\

     
     & \loralap~\cite{yang2023bayesian} &  \textbf{4.16\scriptsize{$\pm$0.58}} & \underline{19.27\scriptsize{$\pm$4.13}} & 27.21\scriptsize{$\pm$4.08} & \underline{7.80\scriptsize{$\pm$0.98}} & 14.73\scriptsize{$\pm$1.78} & \underline{5.76\scriptsize{$\pm$1.59}} \\

    \cmidrule{2-8}

     & \ours~(N=0)  & 14.93\scriptsize{$\pm$0.97} & 20.51\scriptsize{$\pm$0.72} & 9.63\scriptsize{$\pm$0.38} & 13.31\scriptsize{$\pm$0.45} & 9.95\scriptsize{$\pm$0.98} &  7.13\scriptsize{$\pm$0.20}
     \\
     
     & \ours~(N=10)  & \underline{10.43\scriptsize{$\pm$0.71}} & \textbf{11.93\scriptsize{$\pm$0.50}} & \textbf{5.38\scriptsize{$\pm$0.38}} & \textbf{6.82\scriptsize{$\pm$1.29}} & \textbf{3.67\scriptsize{$\pm$0.66}} & \textbf{3.73\scriptsize{$\pm$0.48}}
     \\
     \midrule
    
     \multirow{9}{*}{NLL~($\downarrow$)} & \loramle & 0.82\scriptsize{$\pm$0.03} & 1.48\scriptsize{$\pm$0.10} & 0.86\scriptsize{$\pm$0.07} & 0.88\scriptsize{$\pm$0.01} & 1.15\scriptsize{$\pm$0.06} & 0.67\scriptsize{$\pm$0.01} \\
     
     & \loramap & 0.83\scriptsize{$\pm$0.08} & 1.53\scriptsize{$\pm$0.10} & 0.79\scriptsize{$\pm$0.02} & 0.85\scriptsize{$\pm$0.01} & 0.97\scriptsize{$\pm$0.05} & 0.74\scriptsize{$\pm$0.02} \\
     
     & \loramcd~\cite{gal2016dropout} & 
     0.80\scriptsize{$\pm$0.04}  & 1.53\scriptsize{$\pm$0.03}  & 0.77\scriptsize{$\pm$0.06}  & 0.80\scriptsize{$\pm$0.01}  & 0.97\scriptsize{$\pm$0.09}  & 0.65\scriptsize{$\pm$0.05} \\
     
     & \loraens~\cite{lakshminarayanan2017simple,balabanov2024uncertainty,wang2023lora} & 0.72\scriptsize{$\pm$0.06}  & 1.30\scriptsize{$\pm$0.05}  & 0.58\scriptsize{$\pm$0.00}  & 0.56\scriptsize{$\pm$0.03}  & 0.74\scriptsize{$\pm$0.04}  & 0.46\scriptsize{$\pm$0.01} \\

     
     & \loralap~\cite{yang2023bayesian} & \textbf{0.61\scriptsize{$\pm$0.00}} & 1.46\scriptsize{$\pm$0.02} & 1.41\scriptsize{$\pm$0.02} & 0.55\scriptsize{$\pm$0.00} & 1.04\scriptsize{$\pm$0.01} & 0.47\scriptsize{$\pm$0.01} \\

    \cmidrule{2-8}

     & \ours~(N=0) & 
     0.70\scriptsize{$\pm$0.02} & \underline{1.07\scriptsize{$\pm$0.03}} & \underline{0.56\scriptsize{$\pm$0.01}} & 0.63\scriptsize{$\pm$0.00} & \underline{0.57\scriptsize{$\pm$0.03}} &  \underline{0.39\scriptsize{$\pm$0.01}} 
     \\

     & \ours~(N=10) & 
     \underline{0.65\scriptsize{$\pm$0.01}} & \textbf{0.84\scriptsize{$\pm$0.01}} & \textbf{0.44\scriptsize{$\pm$0.00}} & \textbf{0.54\scriptsize{$\pm$0.01}} & \textbf{0.46\scriptsize{$\pm$0.01}} &  \textbf{0.31\scriptsize{$\pm$0.00}} 
     \\
    \bottomrule[0.12em]
    \end{tabular}
 }
\end{center}
\label{tab:5epochs-llama}
\vspace{-1em}
\end{table*}

\textbf{Additional Details.}
We initialize standard deviation parameterization matrix $\mG$ by element-wise sampling from a uniform distribution with a range of $[\frac{\epsilon}{\sqrt{2}}, \epsilon]$, while keeping the remaining initialization settings consistent with LoRA. 
{To maintain consistency, we use the same learning rate scheduler and warmup ratio for the optimizer of the KL term as we do for the likelihood term.}
We sample only once during the training process. During inference, we sample $N$ times, then take the average of the logits obtained after passing through the softmax function.
Detailed hyperparameter settings are provided in the Table~\ref{table.hyperp-blob}. Table~\ref{table.hyperp-base} provides the hyperparameters for fine-tuning with LoRA shared with other baselines. Our experiments on Llama2-7B were conducted using 2 NVIDIA RTX A5000 GPUs for parallel training, while experiments on RoBERTa-base were conducted using 4 NVIDIA RTX A5000 GPUs for parallel training.

\vspace{1em}

\subsection{Evaluation Metrics for Uncertainty Estimation}
\label{app:implementation-metrics}

Negative Log-Likelihood (\textbf{NLL}) and Expected Calibration Error (\textbf{ECE}~\cite{naeini2015obtaining}) are two prevalent metrics for assessing uncertainty estimation. 
NLL calculates the sum of the negative expected log probability of predicting the actual label. Suppose this predicted probability is given by the model $P_\vtheta$, and we have a test dataset $\{\vx_n, y_n\}_{n=1}^N$ of size $N$. Then the NLL measured on this dataset is 
\begin{equation}
    \text{NLL} = \frac{1}{N}\sum_{n=1}^{N} -\log P_\vtheta(y_n).
\end{equation}

This metric prefers models that assign higher probabilities to correct labels. If the model exhibits overconfident in an incorrect prediction, the probability assigned to the correct label will be diminished, thereby increasing the NLL.

On the other hand, ECE measures how well the model's confidence matches its accuracy. This is done by binning the predictions based on their confidence levels and then computing a weighted average of the absolute difference between accuracy and confidence within each bin:

\begin{equation}
    \text{ECE} = \sum_{m=1}^{M} \frac{|B_m|}{n} \left| \text{acc}(B_m) - \text{conf}(B_m) \right|,
\end{equation}

where \(\text{acc}(B_m)\) and \(\text{conf}(B_m)\) denote the average accuracy and confidence within bin \(B_m\), respectively. These are given by:

\begin{equation}
\begin{aligned}
    \text{acc}(B_m) &= \frac{1}{|B_m|} \sum_{i \in B_m} \mathbf{1}(\hat{y}_i = y_i), \\
    \quad \text{conf}(B_m) &= \frac{1}{|B_m|} \sum_{i \in B_m} P(\hat{y}_i),
\end{aligned}
\end{equation}

where \(|B_m|\) is the number of samples in bin \(m\). We set $|B_m| = 15$ across all experiments.

\subsection{Dataset Details}
\label{app:implementation-dataset}

Table~\ref{tab:dataset} summarizes the size of the training set and the number of labels for each dataset. Table~\ref{tab:prompt} summarizes the prompt templates used for common sense reasoning tasks.

\begin{table*}[h]
\caption{
    Size of the training set and number of labels for each dataset.
}
\begin{center}
\resizebox{1\linewidth}{!}{%
\setlength{\tabcolsep}{3pt}
\begin{tabular}{cccc cccc cccc}
	\toprule[0.12em]
     & WG-S~\cite{wg}
     & ARC-C~\cite{arc} 
     & ARC-E~\cite{arc} 
     & WG-M~\cite{wg} 
     & OBQA~\cite{obqa} 
      
    & RTE~\cite{rte} 
    & MRPC~\cite{mrpc}
    & WiC~\cite{wic}
    & CoLA~\cite{cola}
    & BoolQ~\cite{boolq}\\

     \midrule

     \multirow{1}{*}{Size of Train. Set} 
     & 640 & 
     1.12k & 
     2.25k &  
     2.56k & 
     4.96k & 
     2.49k &
     3.67k &
     5.43k &
     8.55k &
     9.43k \\
         
     \midrule
    
     \multirow{1}{*}{Size of Label Space} 
     &  2 & 
     5 & 
     5 &  
     2 & 
     4 & 
     2 &
     2 &
     2 &
     2 &
     2 \\

    \bottomrule[0.12em]
    \end{tabular}

}
\end{center}
\label{tab:dataset}
\end{table*}

\begin{table}[h!]
\caption{Prompt templates for common sense reasoning tasks.}
\begin{center}
\resizebox{0.76\linewidth}{!}{%
\begin{tabular}{>{\centering\arraybackslash}m{4cm}|>{\centering\arraybackslash}m{9cm}}
\hline
\textbf{Task} & \textbf{Prompt} \\ \hline
Winogrande (WG-S/WG-M) & Select one of the choices that answers the following question: \newline
\{question\} Choices: A. \{option1\}. B. \{option2\}. Answer: \\ \hline
ARC (ARC-C/ARC-E), \newline Openbook QA (OBQA), \newline MMLU  & Select one of the choices that answers the following question: \newline
\{question\} Choices: A. \{choice1\}. B. \{choice2\}. C. \{choice3\}. D. \{choice4\}. Answer: \\ \hline

BoolQ & Answer the question with only True or False: \newline
\{question\} Context: \{passage\}. \\ 
\hline
\end{tabular}
}
\end{center}
\label{tab:prompt}
\end{table}

\section{Additional Experimental Results}
\label{app:add-exp}
This section provides additional experimental results omitted from the main body of the paper due to space limitations. First, we present the results of \ours when applied to RoBERTa, another pre-trained language model, in \appref{app:add-exp-robert}. 
Next, in \appref{sec:ablation}, we conduct the ablation study on our proposed refinement in \ours. 
Then we analyze the memory and training time costs in \appref{sec:exp-mem cost}.
Finally, we provide visualization illustrating our \ours's advantage on embedding uncertainty in \appref{app:visual}.

\subsection{Performance of RoBERTa on In-distribution Datasets}
\label{app:add-exp-robert}

\begin{table*}[t]
\caption{
    \textbf{Performance of different methods applied to LoRA on RoBERTa-base pre-trained weights.} The evaluation is undertaken on five GLUE~\cite{glue} and SuperGLUE~\cite{superglue} tasks, with a shared hyper-parameter setting without using individual validation dataset. ``$\uparrow$'' and ``$\downarrow$'' represent that higher and lower values are preferred, respectively. 
    The \textbf{boldface} and \underline{underline} are used to denote the best and runner-up performance, respectively. 
    The asterisk ``$^\ast$'' denotes training failure.
}
\begin{center}
\resizebox{1\linewidth}{!}{%
\setlength{\tabcolsep}{10pt}
\begin{tabular}{clccc ccc}
	\toprule[0.12em]
	\multirow{2}{*}[-0.25em]{\textbf{Metric}} & \multirow{2}{*}[-0.25em]{\textbf{Method}} & \multicolumn{5}{c}{\textbf{Datasets}}
     \\
     \cmidrule{3-7}
    & & RTE~\cite{rte} 
    & MRPC~\cite{mrpc}
    & WiC~\cite{wic}
    & CoLA~\cite{cola} 
    & BoolQ~\cite{boolq} \\
     
     \midrule

     \multirow{8}{*}{ACC~($\uparrow$)} & \loramle & 75.81\scriptsize{$\pm$0.78} & 86.27\scriptsize{$\pm$0.69} & 64.52\scriptsize{$\pm$0.91} &  83.29\scriptsize{$\pm$0.16} & 77.67\scriptsize{$\pm$0.51} \\
     
     & \loramap & 75.81\scriptsize{$\pm$2.26} & 86.36\scriptsize{$\pm$0.51} & 65.46\scriptsize{$\pm$1.04} & 83.00\scriptsize{$\pm$0.15} & 77.69\scriptsize{$\pm$0.65} \\
     
     & \loramcd~\cite{gal2016dropout} & 76.65\scriptsize{$\pm$0.85} & 87.75\scriptsize{$\pm$0.53} & 68.55\scriptsize{$\pm$0.32} & 84.76\scriptsize{$\pm$0.62} & 78.41\scriptsize{$\pm$0.25}  \\
     
     & \loraens~\cite{lakshminarayanan2017simple,balabanov2024uncertainty,wang2023lora} & 77.74\scriptsize{$\pm$1.10} & 88.64\scriptsize{$\pm$0.37} & 65.83\scriptsize{$\pm$0.41} & 84.08\scriptsize{$\pm$0.44} & 78.57\scriptsize{$\pm$0.36} \\

     & BBB~\cite{blundell2015BBB} 
     & 49.46$^\ast$\scriptsize{$\pm$2.53}
     & 68.38\scriptsize{$\pm$0.00}
     & 50.57$^\ast$\scriptsize{$\pm$1.74}
     & 69.13\scriptsize{$\pm$0.00}
     & 62.16\scriptsize{$\pm$0.04}
     \\
     
     & \loralap~\cite{yang2023bayesian} & 76.05\scriptsize{$\pm$0.95} & 86.52\scriptsize{$\pm$0.72} & 64.52\scriptsize{$\pm$0.91} & 83.29\scriptsize{$\pm$0.16} & 77.67\scriptsize{$\pm$0.52} \\
     

     \cmidrule{2-7}

     & \ours~(N=0) 
     & 76.05\scriptsize{$\pm$0.17}
     & 88.24\scriptsize{$\pm$0.00}
     & 63.17\scriptsize{$\pm$0.22}
     & 80.92\scriptsize{$\pm$0.70}
     & 74.80\scriptsize{$\pm$2.10}
     \\

     & \ours~(N=5) 
     & 74.61\scriptsize{$\pm$0.61}
     & 88.48\scriptsize{$\pm$0.60}
     & 64.00\scriptsize{$\pm$0.53}
     & 80.54\scriptsize{$\pm$0.16}
     & 74.77\scriptsize{$\pm$1.77}
     \\

     & \ours~(N=10) 
     & 75.45\scriptsize{$\pm$0.51}
     & 88.73\scriptsize{$\pm$0.35}
     & 64.26\scriptsize{$\pm$1.00}
     & 80.89\scriptsize{$\pm$0.24}
     & 75.49\scriptsize{$\pm$1.60}
     \\
     
     \midrule

     \multirow{8}{*}{ECE~($\downarrow$)} & \loramle & 20.59\scriptsize{$\pm$1.25} & 11.13\scriptsize{$\pm$1.05} & 25.72\scriptsize{$\pm$0.83} &  10.70\scriptsize{$\pm$0.49} & 10.02\scriptsize{$\pm$0.71} \\
     
     & \loramap & 21.67\scriptsize{$\pm$3.25} & 11.12\scriptsize{$\pm$0.45} & 24.26\scriptsize{$\pm$1.17} & 10.61\scriptsize{$\pm$0.49} & 10.11\scriptsize{$\pm$0.62} \\
     
     & \loramcd~\cite{gal2016dropout} & 13.06\scriptsize{$\pm$0.59} & 7.36\scriptsize{$\pm$0.85} & 16.94\scriptsize{$\pm$0.75} & \underline{4.58\scriptsize{$\pm$0.27}} & 6.21\scriptsize{$\pm$0.33}  \\
     
     & \loraens~\cite{lakshminarayanan2017simple,balabanov2024uncertainty,wang2023lora} & 19.47\scriptsize{$\pm$0.37} & 10.13\scriptsize{$\pm$0.56} & 28.62\scriptsize{$\pm$0.63} & 12.44\scriptsize{$\pm$0.42} & 5.98\scriptsize{$\pm$0.26} \\

     & BBB~\cite{blundell2015BBB} 
     & 2.66$^\ast$\scriptsize{$\pm$2.24}
     & 6.46\scriptsize{$\pm$0.43}
     & 2.53$^\ast$\scriptsize{$\pm$0.39}
     & 3.90$^*$\scriptsize{$\pm$0.41}
     & 5.02$^*$\scriptsize{$\pm$0.29}
     \\
     
     & \loralap~\cite{yang2023bayesian} & \textbf{5.33\scriptsize{$\pm$0.60}} & 6.29\scriptsize{$\pm$0.99} & \textbf{11.48\scriptsize{$\pm$0.67}} & \textbf{3.13\scriptsize{$\pm$0.28}} & 4.84\scriptsize{$\pm$0.15} \\

     \cmidrule{2-7}

     & \ours~(N=0) 
     & 14.64\scriptsize{$\pm$0.75}
     & 5.61\scriptsize{$\pm$0.06}
     & 18.93\scriptsize{$\pm$1.39}
     & 10.90\scriptsize{$\pm$0.24}
     & 5.80\scriptsize{$\pm$0.41}
     \\

     & \ours~(N=5) 
     & 10.46\scriptsize{$\pm$0.61}
     & \underline{4.49\scriptsize{$\pm$0.32}}
     & 13.62\scriptsize{$\pm$1.18}
     & 7.76\scriptsize{$\pm$0.21}
     & \underline{3.21\scriptsize{$\pm$0.13}}
     \\

     & \ours~(N=10) 
     & \underline{8.97\scriptsize{$\pm$0.98}}
     & \textbf{3.30\scriptsize{$\pm$0.19}}
     & \underline{13.03\scriptsize{$\pm$0.85}}
     & 7.83\scriptsize{$\pm$0.27}
     & \textbf{2.90\scriptsize{$\pm$0.12}}
     \\

     \midrule

     \multirow{8}{*}{NLL~($\downarrow$)} & \loramle & 1.11\scriptsize{$\pm$0.02} & 0.62\scriptsize{$\pm$0.02} & 1.19\scriptsize{$\pm$0.03} &  0.53\scriptsize{$\pm$0.02} & 0.56\scriptsize{$\pm$0.01} \\
     
     & \loramap & 1.23\scriptsize{$\pm$0.10} & 0.58\scriptsize{$\pm$0.04} & 1.14\scriptsize{$\pm$0.06} & 0.53\scriptsize{$\pm$0.00} & 0.55\scriptsize{$\pm$0.02} \\
     
     & \loramcd~\cite{gal2016dropout} & 0.65\scriptsize{$\pm$0.03} & 0.39\scriptsize{$\pm$0.04} & 0.88\scriptsize{$\pm$0.02} & \textbf{0.39\scriptsize{$\pm$0.01}} & \underline{0.50\scriptsize{$\pm$0.01}} \\
     
     & \loraens~\cite{lakshminarayanan2017simple,balabanov2024uncertainty,wang2023lora} & 1.04\scriptsize{$\pm$0.05} & 0.63\scriptsize{$\pm$0.02} & 1.70\scriptsize{$\pm$0.07} & 0.62\scriptsize{$\pm$0.00} & \textbf{0.48\scriptsize{$\pm$0.00}} \\

     & BBB~\cite{blundell2015BBB} 
     & 0.69$^\ast$\scriptsize{$\pm$0.00}
     & 0.63\scriptsize{$\pm$0.00}
     & 0.69$^\ast$\scriptsize{$\pm$0.00}
     & 0.62\scriptsize{$\pm$0.00}
     & 0.67\scriptsize{$\pm$0.00}
     \\

     & \loralap~\cite{yang2023bayesian} & 0.55\scriptsize{$\pm$0.00} & 0.47\scriptsize{$\pm$0.01} & \textbf{0.63\scriptsize{$\pm$0.00}} & 0.48\scriptsize{$\pm$0.00} & 0.53\scriptsize{$\pm$0.00} \\
     

    \cmidrule{2-7}

    & \ours~(N=0) 
     & 0.56\scriptsize{$\pm$0.01}
     & 0.29\scriptsize{$\pm$0.00}
     & 0.76\scriptsize{$\pm$0.02}
     & 0.52\scriptsize{$\pm$0.01}
     & 0.52\scriptsize{$\pm$0.02}
     \\

    & \ours~(N=5) 
     & \underline{0.50\scriptsize{$\pm$0.01}}
     & \underline{0.27\scriptsize{$\pm$0.00}}
     & 0.68\scriptsize{$\pm$0.01}
     & \underline{0.45\scriptsize{$\pm$0.01}}
     & 0.51\scriptsize{$\pm$0.02}
     \\

    & \ours~(N=10) 
     & \textbf{0.48\scriptsize{$\pm$0.01}}
     & \textbf{0.26\scriptsize{$\pm$0.00}}
     & \underline{0.67\scriptsize{$\pm$0.01}}
     & 0.46\scriptsize{$\pm$0.01}
     & 0.51\scriptsize{$\pm$0.01}
     \\
     
	\bottomrule[0.12em]
	\end{tabular}
	}
\end{center}
\label{tab:main-roberta-base}
\end{table*}

We also evaluate different methods on RoBERTa-base, which has approximately $\nicefrac{1}{50}$ the parameter count of Llama2-7B. Table~\ref{tab:main-roberta-base} shows the results. Compared to MLE, MAP shows minor improvements in NLL and ECE, while MCD, ENS, and LAP enjoy significant improvements. The convergence difficulty observed with the BBB algorithm is further exacerbated on the smaller model, resulting in significant decreases in ACC across all datasets, and even training failures on RTE and WiC. In contrast, our method demonstrates the best or runner-up performance in uncertainty estimation on almost all datasets. Only a slight decrease in ACC is observed on BoolQ and CoLA. 
We suspect that such decrease is caused by RoBERTa-base's small model size compared to the large size of these datasets BoolQ and CoLA (i.e., underfitting). Using a larger pretrained model, e.g., Llama2-7B, would potentially address this issue. 


\subsection{Ablation Study}
\label{sec:ablation}

We perform an ablation study on the Llama2-7B model to showcase the effects of a range of techniques we designed: KL {R}e-{W}eighting~(RW, \appref{kl-reweighting}), 
{R}e-{P}arameterization~(RP, \Secref{sec:method-reparam}), and 
{A}symmetric {B}ayesianization~(AB, \Secref{sec:method-posterior}). 
In the scenarios w/o AB, we Bayesianize both matrices, $\mA$ and $\mB$.
In practice, using identical initialization and prior for the standard deviation matrix $\mG$ of the variational distribution on both $\mA$ and $\mB$ leads to training failures caused by ``NaN'' loss across all datasets; this is consistent with the findings in \Secref{sec:method-posterior}. 
As a solution, we introduce a scaled standard deviation matrix $\nicefrac{\mG}{100}$ on $\mB$ to alleviate early-stage fluctuations. 
Nevertheless, it is important to note that not using AB incurs double the additional memory cost and training time, as described in Appendix~\ref{sec:exp-mem cost}.

As demonstrated in Table~\ref{tab:ablation-llama},  BBB w/o AB fails to converge due to the unbounded NaN loss, which cannot be solved by using scaled standard deviation.
By introducing KL Re-Weighting, Re-Parameterization, and scaled standard deviation, \ours w/o AB achieves the runner-up performance and improves accuracy on small datasets. 
However, \ours with all techniques achieves the best ECE and NLL with minimal additional computational cost.

\begin{table*}[t]
\caption{
    \textbf{Ablation study of \ours, applied to LoRA on Llama2-7B pre-trained weights,} where \textbf{RW}, \textbf{RP}, and \textbf{AB} represent our designed techniques of 
    KL \textbf{R}e-\textbf{W}eighting~(\appref{kl-reweighting}), 
    \textbf{R}e-\textbf{P}arameterization~(\Secref{sec:method-reparam}), and 
    \textbf{A}symmetric \textbf{B}ayesianization~(\Secref{sec:method-posterior}), respectively. 
    The evaluation is done following \Tabref{tab:main-llama}.
    We set the number of samples during training $K=1$ and the number of samples during inference $N=10$ across the variants~(denoted by ``$-$'') of the BBB~\cite{blundell2015BBB} and \ours for fair comparison. 
    {We denote by ``$^\ast$'' experiments with the scaled standard deviation matrix. 
    The hyphen ``$-$'' in the table denotes training failure caused by ``NaN'' loss.}
    ``$\uparrow$'' and ``$\downarrow$'' indicate that higher and lower values are preferred, respectively. 
    \textbf{Boldface} and \underline{underlining} denote the best and the second-best performance, respectively. 
}
\begin{center}
\resizebox{1\linewidth}{!}{%
\setlength{\tabcolsep}{5pt}

\begin{tabular}{cl ccc cccccc}
	\toprule[0.12em]
	\multirow{2}{*}[-0.25em]{\textbf{Metric}} & \multirow{2}{*}[-0.25em]{\textbf{Method}} & \multicolumn{3}{c}{\textbf{Techniques}} & \multicolumn{6}{c}{\textbf{Datasets}}
     \\
     \cmidrule(lr){3-5}\cmidrule(lr){6-11}
     & & RW & RP & AB & WG-S~\cite{wg}
     & ARC-C~\cite{arc} 
     & ARC-E~\cite{arc} 
     & WG-M~\cite{wg} 
     & OBQA~\cite{obqa} 
     & BoolQ~\cite{boolq} \\
     \midrule

     \multirow{9}{*}{ACC~($\uparrow$)} & \loramle & - & - & - & 68.99\scriptsize{$\pm$0.58} & \underline{69.10\scriptsize{$\pm$2.84}} & 85.65\scriptsize{$\pm$0.92} & 74.53\scriptsize{$\pm$0.66} & 81.52\scriptsize{$\pm$0.25} & 86.53\scriptsize{$\pm$0.28} \\

     \cmidrule{2-11}

     &  \loranaivebbb- & & & & 
     - &
     - &
     - &
     - &
     - &
     - 
     \\
     
     &  \loranaivebbb-$^*$ & & & & 
     - &
     - &
     - &
     - &
     - &
     - 
     \\
     
     &  \loranaivebbb~\cite{blundell2015BBB} & & & \cmark & 
     56.54\scriptsize{$\pm$7.87} & 
     68.13\scriptsize{$\pm$1.27} & 
     85.86\scriptsize{$\pm$0.74} & 
     73.63\scriptsize{$\pm$2.44} & 
     \textbf{82.06\scriptsize{$\pm$0.59}} & 
     \underline{87.21\scriptsize{$\pm$0.22}} 
     \\
     
     \cmidrule{2-11} 
    
     & \ours- & \cmark & & \cmark &
     \textbf{69.75\scriptsize{$\pm$0.60}} &
     67.91\scriptsize{$\pm$1.43} &
     \underline{86.03\scriptsize{$\pm$0.74}} &
     \textbf{76.24\scriptsize{$\pm$0.55}} &
     \underline{81.65\scriptsize{$\pm$0.66}} &
     \textbf{87.23\scriptsize{$\pm$0.42}} 
     \\

     & \ours- & & \cmark & \cmark &
     - &
     - &
     - &
     - &
     - &
     - 
     \\

     & \ours- & \cmark & \cmark &   &
     - &
     - &
     - &
     - &
     - &
     - 
     \\

     & \ours-$^*$ & \cmark & \cmark &   &
      \textbf{69.75\scriptsize{$\pm$0.26}} &
      \textbf{70.27\scriptsize{$\pm$0.48}} &
      \textbf{86.33\scriptsize{$\pm$0.44}} &
      \underline{74.92\scriptsize{$\pm$0.19}} &
      81.32\scriptsize{$\pm$0.41} &
      86.47\scriptsize{$\pm$0.46}
     \\
     
     & \ours~(Ours) & \cmark & \cmark & \cmark & 
     \underline{69.07\scriptsize{$\pm$0.34}} & 
     68.81\scriptsize{$\pm$1.09} & 
     85.56\scriptsize{$\pm$0.35} & 
     73.69\scriptsize{$\pm$0.17} & 
     81.52\scriptsize{$\pm$0.74} & 
     86.99\scriptsize{$\pm$0.24} 
     \\
     \midrule
    
     \multirow{9}{*}{ECE~($\downarrow$)} & \loramle & - & - & - & 29.83\scriptsize{$\pm$0.58} & 29.00\scriptsize{$\pm$1.97} & 13.12\scriptsize{$\pm$1.39} & 20.62\scriptsize{$\pm$0.74} & 12.55\scriptsize{$\pm$0.46} & 3.18\scriptsize{$\pm$0.09} \\

     \cmidrule{2-11}

     &  \loranaivebbb- & & & & 
     - &
     - &
     - &
     - &
     - &
     - 
     \\
     
     &  \loranaivebbb-$^*$ & & & & 
     - &
     - &
     - &
     - &
     - &
     - 
     \\
     
     &  \loranaivebbb~\cite{blundell2015BBB} & & & \cmark &  
     21.81\scriptsize{$\pm$12.95} & 
     26.23\scriptsize{$\pm$1.47} & 
     12.28\scriptsize{$\pm$0.58} & 
     15.76\scriptsize{$\pm$4.71} & 
     11.38\scriptsize{$\pm$1.07} & 
     3.74\scriptsize{$\pm$0.10}
     \\
     
    \cmidrule{2-11}

     & \ours- & \cmark & & \cmark &
     26.60\scriptsize{$\pm$0.78} &
     26.24\scriptsize{$\pm$0.94} &
     11.53\scriptsize{$\pm$0.52} &
     18.05\scriptsize{$\pm$0.76} &
     12.36\scriptsize{$\pm$0.42} &
     3.05\scriptsize{$\pm$0.09} 
     \\

     & \ours- & & \cmark & \cmark &
     - &
     - &
     - &
     - &
     - &
     - 
     \\

     & \ours- & \cmark & \cmark &   &
     - &
     - &
     - &
     - &
     - &
     - 
     \\

     & \ours-$^*$ & \cmark & \cmark &   &
       \underline{16.59\scriptsize{$\pm$0.57}} &
       \underline{13.85\scriptsize{$\pm$1.06}} &
       \underline{5.93\scriptsize{$\pm$0.63}} &
       \underline{8.33\scriptsize{$\pm$0.78}} &
       \underline{4.77\scriptsize{$\pm$0.26}} &
       \textbf{1.18\scriptsize{$\pm$0.20}}
     \\
     
     & \ours~(Ours) & \cmark & \cmark & \cmark &
     \textbf{9.35\scriptsize{$\pm$1.37}} & 
     \textbf{9.59\scriptsize{$\pm$1.88}} & 
     \textbf{3.64\scriptsize{$\pm$0.53}} & 
     \textbf{3.01\scriptsize{$\pm$0.12}} & 
     \textbf{3.77\scriptsize{$\pm$1.47}} & \underline{1.41\scriptsize{$\pm$0.19}} 
     \\
     \midrule
    
     \multirow{9}{*}{NLL~($\downarrow$)} & \loramle & - & - & - & 3.17\scriptsize{$\pm$0.37} & 2.85\scriptsize{$\pm$0.27} & 1.17\scriptsize{$\pm$0.13} & 0.95\scriptsize{$\pm$0.07} & 0.73\scriptsize{$\pm$0.03} & \underline{0.32\scriptsize{$\pm$0.00}} \\

     \cmidrule{2-11}

     &  \loranaivebbb- & & & & 
     - &
     - &
     - &
     - &
     - &
     - 
     \\
     
     &  \loranaivebbb-$^*$ & & & & 
     - &
     - &
     - &
     - &
     - &
     - 
     \\
     
     & \loranaivebbb~\cite{blundell2015BBB} & & & \cmark &
     1.40\scriptsize{$\pm$0.55} & 
     2.23\scriptsize{$\pm$0.04} & 
     0.91\scriptsize{$\pm$0.06} & 
     0.84\scriptsize{$\pm$0.15} & 
    0.66\scriptsize{$\pm$0.05} & 
     \textbf{0.31\scriptsize{$\pm$0.00}}
     \\

    \cmidrule{2-11}

     & \ours- & \cmark & & \cmark &
     1.96\scriptsize{$\pm$0.20} &
     2.31\scriptsize{$\pm$0.13} &
     0.84\scriptsize{$\pm$0.03} &
     0.87\scriptsize{$\pm$0.01} &
     0.68\scriptsize{$\pm$0.00} &
     \textbf{0.31\scriptsize{$\pm$0.00}} 
     \\

     & \ours- & & \cmark & \cmark &
     - &
     - &
     - &
     - &
     - &
     - 
     \\

     & \ours- & \cmark & \cmark &   &
     - &
     - &
     - &
     - &
     - &
     - 
     \\

     & \ours-$^*$ & \cmark & \cmark &   &
     \underline{0.80\scriptsize{$\pm$0.02}} &
     \underline{0.91\scriptsize{$\pm$0.04}} &
     \underline{0.46\scriptsize{$\pm$0.01}} &
     \underline{0.55\scriptsize{$\pm$0.01}} &
     \underline{0.51\scriptsize{$\pm$0.00}} &
     \underline{0.32\scriptsize{$\pm$0.00}} 
     \\
     
     & \ours~(Ours) & \cmark & \cmark & \cmark & 
     \textbf{0.63\scriptsize{$\pm$0.01}} & 
     \textbf{0.78\scriptsize{$\pm$0.02}} & 
     \textbf{0.40\scriptsize{$\pm$0.01}} & 
     \textbf{0.54\scriptsize{$\pm$0.00}} & 
    \textbf{0.50\scriptsize{$\pm$0.01}} & \textbf{0.31\scriptsize{$\pm$0.00}} 
     \\
    \bottomrule[0.12em]
    \end{tabular}
 }
\end{center}
\label{tab:ablation-llama}
\vspace{-1em}
\end{table*}

\subsection{{Additional Results on Memory and Time Efficiency}}
\label{sec:exp-mem cost}

By introducing an additional standard deviation matrix $\mOmega$ of the same size as the LoRA $\mA$ matrix, the number of trainable parameters in \ours increases by half compared to LoRA. 
{In the case of \ours w/o Asymmetric Bayesianization~(AB), the number of trainable parameters are twice as many as those in LoRA.}
The calculation of KL divergence and the inclusion of the additional standard deviation matrix in the likelihood loss computation result in additional forward and backward propagation time. 
We conduct parallel training using two NVIDIA RTX A5000 GPUs to observe the differences in GPU memory cost and training time between \ours and standard LoRA fine-tuning on the Llama2-7B model. 
The results are shown in Table~\ref{tab:mem_cost}. 
{\ours increases memory cost by only about 3\% to 13\% compared to LoRA, with training time increased by about 15\%. }

{We further evaluate the inference time and maximum memory usage for standard LoRA, Laplace-LoRA~(LAP), and our \ours, as shown in \Tabref{tab:mem_cost_inf}. {The experiments are conducted on two NVIDIA A100 GPUs.}
These results show that compared to LAP, our BLoB can achieve comparable or better ECE and NLL with less inference time and less memory usage. Notably, our BLoB's memory overhead compared to standard LoRA is minimal, while LAP introduces significant memory overhead.}

\begin{table*}[!t]
\caption{
    \textbf{A comparison of time and maximum memory cost between standard LoRA and BLoB, during training.} The evaluation is based on fine-tuning for 5,000 steps on the Llama2-7B model.
}
\begin{center}
\resizebox{1\linewidth}{!}{%
\setlength{\tabcolsep}{6pt}
\begin{tabular}{clccc cccc}
	\toprule[0.12em]
	\multirow{2}{*}[-0.25em]{\textbf{Metric}} & \multirow{2}{*}[-0.25em]{\textbf{Method}} & \multicolumn{6}{c}{\textbf{Datasets}}
     \\
     \cmidrule{3-8}
     & & WG-S~\cite{wg}
     & ARC-C~\cite{arc} 
     & ARC-E~\cite{arc} 
     & WG-M~\cite{wg} 
     & OBQA~\cite{obqa} 
     & BoolQ~\cite{boolq} \\
     \midrule

     \multirow{2}{*}{Time (Seconds)~($\downarrow$)} 
     & Standard LoRA & 
     1399 & 
     1614 & 
     1586 &  
     1408 & 
     1822 & 
     3382 \\

     & \ours (N=10) & 
     1563 & 
     1790 & 
     1753 &
     1556 & 
     2142 & 
     3733 \\

     \midrule
    
     \multirow{2}{*}{Max Memory (MB)~($\downarrow$)} 
     & Standard LoRA & 
     14688 & 
     16870 & 
     17044 &  
     14710 & 
     14984 & 
     20784 \\

     & \ours (N=10) & 
     15015  & 
     18863 &
     19015 &
     15015 & 
     15890 & 
     23552 \\
    \bottomrule[0.12em]
    \end{tabular}

}
\end{center}
\label{tab:mem_cost}
\vspace{0em}
\end{table*}

\begin{table*}[h]
\caption{
    \textbf{A comparison of time and max memory cost between Standard LoRA, LAP, and BLoB~(N=10) during inference.} 
}
\begin{center}
\resizebox{1\linewidth}{!}{%
\setlength{\tabcolsep}{7pt}
\begin{tabular}{clcccccc}
    \toprule[0.12em]
    \multirow{2}{*}[-0.25em]{\textbf{Metric}} & \multirow{2}{*}[-0.25em]{\textbf{Method}} & \multicolumn{6}{c}{\textbf{Datasets}} \\
    \cmidrule{3-8}
     & & WG-S~\cite{wg} & ARC-C~\cite{arc} & ARC-E~\cite{arc} & WG-M~\cite{wg} & OBQA~\cite{obqa} & BoolQ~\cite{boolq} \\
    \midrule
    \multirow{3}{*}{Time (Seconds) ($\downarrow$)} 
     & Standard LoRA & 17 & 5 & 8 & 17 & 7 & 58 \\
     & LAP & 311 & 445 & 814 & 554 & 1165 & 2508 \\
     & BLoB (N=10) & 193 & 45 & 86 & 193 & 75 & 627 \\
    \midrule
    \multirow{3}{*}{Max Memory (MB) ($\downarrow$)} 
     & Standard LoRA & 14391 & 14157 & 13081 & 14391 & 14391 & 14160 \\
     & LAP & 43742 & 61881 & 64737 & 43678 & 55642 & 67364 \\
     & BLoB (N=10) & 14171 & 14411 & 13911 & 14407 & 14478 & 14177 \\
    \bottomrule[0.12em]
\end{tabular}
}
\end{center}
\label{tab:mem_cost_inf}
\vspace{0em}
\end{table*}

\subsection{{Impact of Sample Size on Inference Performance}}\label{app:sample-size}

\begin{figure}[h]
    \centering
        \includegraphics[width=\linewidth]{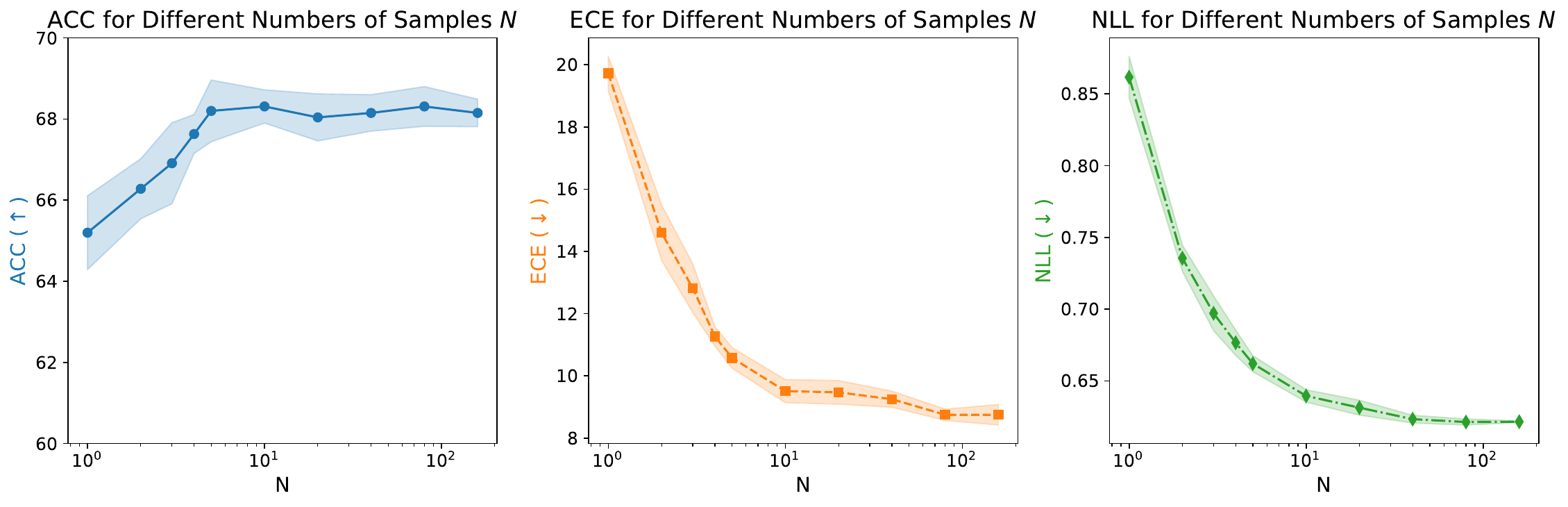}
    \caption{\textbf{Performance of \ours with Varying Sample Sizes $N$ during Inference.} We fine-tune the Llama2-7B model on the WG-S dataset for 5,000 steps, evaluating the model's performance with different sample sizes, specifically when $N$ is 1, 2, 3, 4, 5, 10, 20, 40, 80, and 160.}
    \label{fig:visual_sampleN}
\end{figure}

Here we provide a more detailed empirical study on the sample size of \ours during inference. 
Specifically, we report the results for different number of samples from $N=1$ to $N=160$ on the WG-S dataset, demonstrating improved uncertainty estimation with increased number of samples, as shown in \Figref{fig:visual_sampleN}.

\subsection{{Trade-Off between Accuracy and Calibration Controlled by Gaussian Prior}}\label{app:tradeoff}

\begin{figure}[h]
    \centering
        \includegraphics[width=\linewidth]{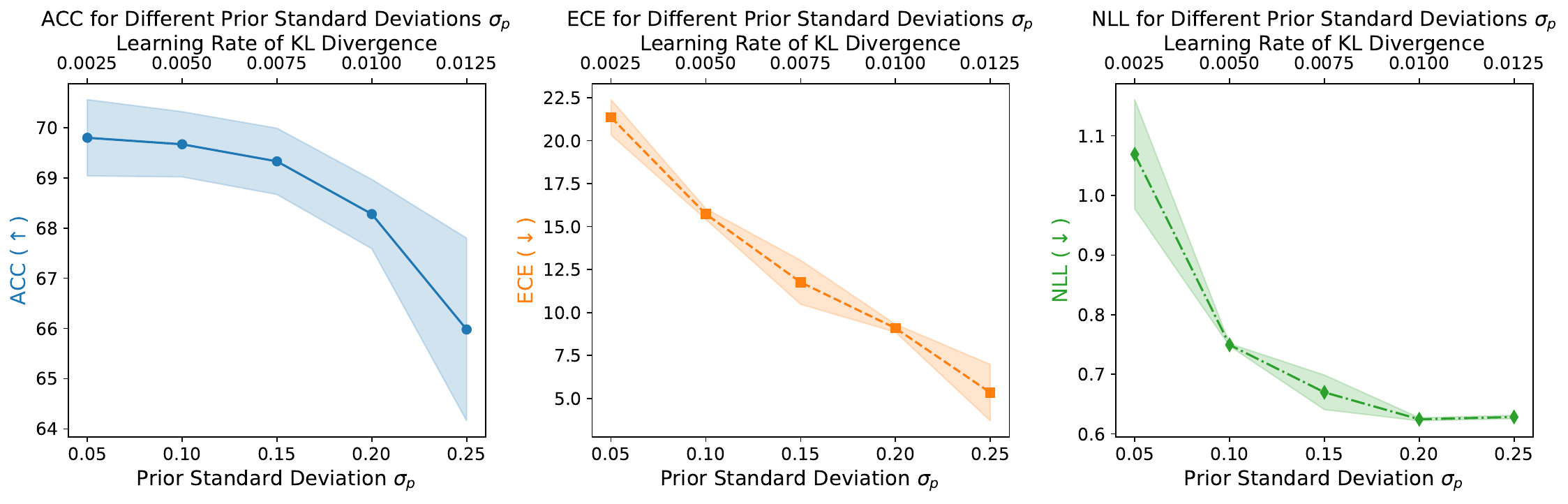}
    \caption{\textbf{Performance of \ours (N=10) with Varying Prior Gaussian Standard Deviations $\sigma_p$.} We fine-tune the Llama2-7B model on the WG-S dataset for 5,000 gradient steps, evaluating the model's performance with different prior Gaussian standard deviations and learning rates of KL divergence.}
    \label{fig:visual_tradeoff}
\end{figure}

{
The empirical trade-off between accuracy and calibration caused by different model architectures is observed in \cite{stengel-eskin-van-durme-2023-calibrated}. By controlling the standard deviation of the prior Gaussian distribution, we observed a similar trade-off between accuracy and calibration. Specifically, we report results for different prior Gaussian standard deviations, ranging from $0.05$ to $0.25$, while proportionally scaling the learning rate of KL divergence from its original value of $0.01$ to values between $0.0025$ and $0.0125$. This highlights the trade-off between accuracy and calibration, as shown in \Figref{fig:visual_tradeoff}.
}

\subsection{Embedding Uncertainty of \ours: A Preliminary Visual Study}
\label{app:visual}
\begin{figure}[t]
    \centering
        \includegraphics[width=\linewidth]{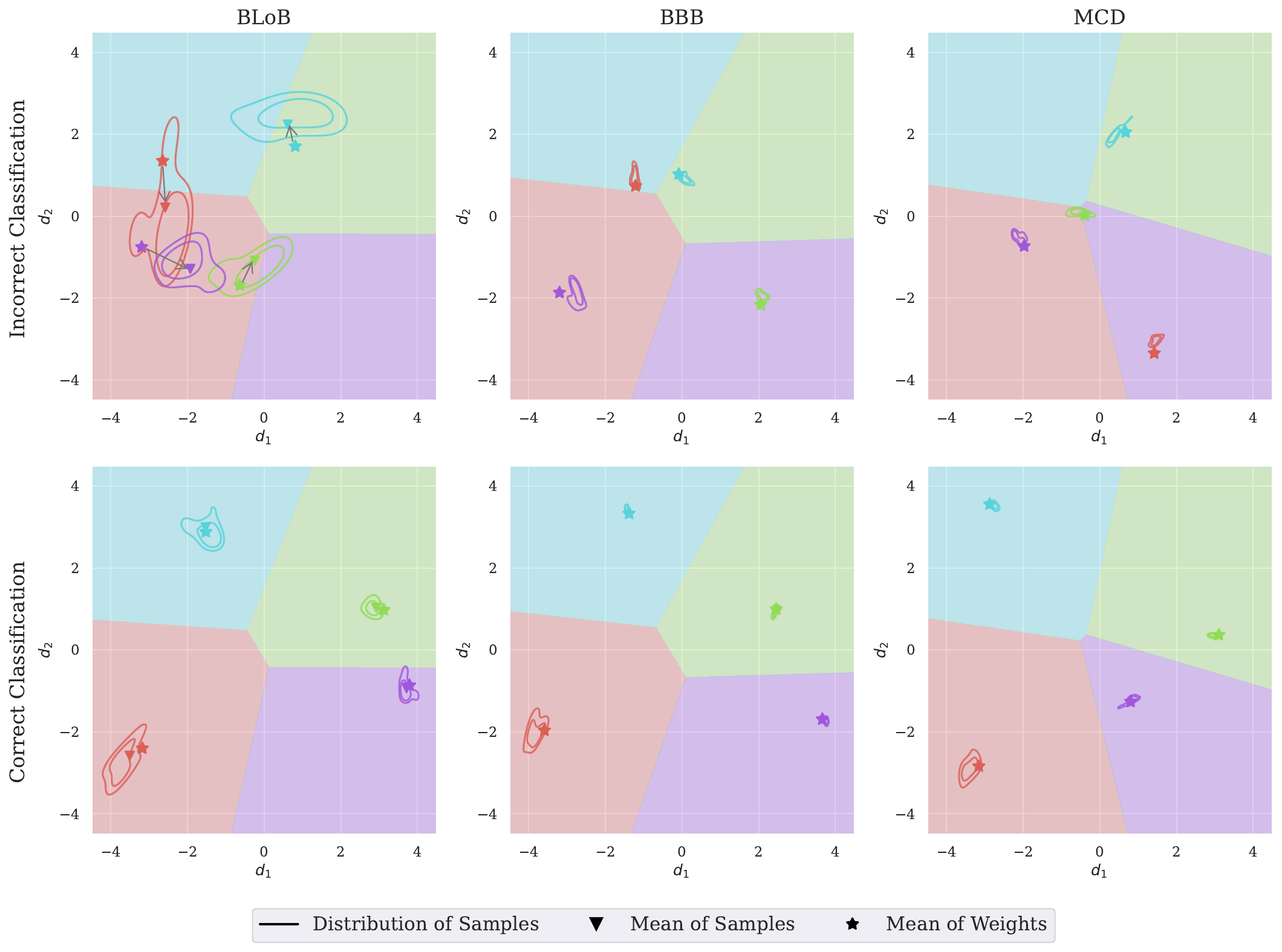}
    \caption{\textbf{Visualization of embedding uncertainty quality for different methods.} The model is fine-tuned for 5,000 steps on the Llama2-7B. We fine-tune the Llama2-7B model on the OBQA dataset for 5000 steps. The two contour lines represent the probability mass of 0.5 and 0.75, respectively.}
    \label{fig:visual}
\end{figure}

Estimating the uncertainty of LLMs in the embedding space has recently garnered significant attention in the community~\cite{chan2023adaptive}. Expressing models' uncertainty via their generated embeddings can benefit both discriminative (the focus of this paper) and generative models. In this section, we present a preliminary study on uncertainty estimation in the embedding spaces of different models, as illustrated in \Figref{fig:visual}. 
We compare \ours with two baseline models, \loranaivebbb and \loramcd, which can generate embedding samples and effectively estimate uncertainty. We exclude \loralap from this section due to its excessive memory consumption, which consistently results in Out-Of-Memory (OOM) errors during inference.
The experiment is conducted on the OBQA dataset~\cite{obqa}, which consists of four categories.


For each input sequence $\vs$, we use the last token's embedding generated by the final transformer block in Llama2-7B as the final embedding. Given the weights $\mW$, we denote the embedding as $\vphi (\vs; \mW)$. 
Generally, three types of embeddings can be generated using the Bayesian approach:
\begin{itemize}[leftmargin=18pt]
    \item[(a)] Embeddings generated by the mean of the weights (these embeddings are shown as ``\starmark'' in \Figref{fig:visual}):
    \begin{align}
    \vphi (\vs; \E_{\mW\sim q(\cdot|\vtheta)}[{\mW}]) 
    &= \vphi (\vs; \mW_0+\mB \mM) .\label{eq:emb_of_mean}
    \end{align}
    \item[(b)] Embedding samples generated by sampling different weights from the approximate posterior, whose distribution is plotted by the solid line (\textbf{---}).
    \item[(c)] The expectation of the embedding, which is approximated by averaging the sampled embeddings:
    \begin{align} 
    \E_{\mW\sim q(\cdot|\vtheta)}[\vphi (\vs; \mW)]
    &\approx \frac{1}{N}\sum_{n=1}^{N} \vphi (\vs; \mW_0 + \mB(\mM + \mE_{n} \circ \mOmega)),\label{eq:mean_of_emb}
    \end{align}
    where $N$ denotes the number of samples during inference, and $\mE_n$ denotes the $n$-th sampled noise for the weight matrix.
    We show this expectation as ``\trianglemark'' in \Figref{fig:visual}.
\end{itemize}

To visually demonstrate the confidence calibration effect of the Bayesian treatment, we adopt the following pipeline of visualization, which we believe can be further applied in visualizing other frameworks' embedding uncertainty quality.
\begin{enumerate}[label=(\arabic*),leftmargin=18pt]

    \item Acquire high-dimensional embeddings produced by the weight mean for the given test dataset, as decribed in (a) and~\eqnref{eq:emb_of_mean} above.
    
    \item Use Linear Discriminant Analysis~(LDA)~\cite{bishop2006pattern} to project these high-dimensional embeddings into a low-dimensional 2D space.
    
    \item In the 2D space, fit a logistic regression model to mimic the decision regions and color them based on the true labels.
    
    \item Sample weights 10 times from the approximate posterior, generate the embeddings, and project them into the same 2D space using the previously learned LDA. Use Kernel Density Estimation (KDE)~\cite{kde1, kde2} to show their distributions, as described in (b) above.
    
    \item Average the sampled embeddings for each example and visualize them in the 2D space, as described in (c) and~\eqnref{eq:mean_of_emb}  above.
    
\end{enumerate}

In \Figref{fig:visual}, we show 4 correct and incorrect predictions made by each model.
Ideally, a model with better uncertainty estimation should produce lower level of uncertainty~(\textbf{smaller embedding variance}, i.e., smaller contours, and \textbf{further away from the decision boundary}) for correct predictions, and higher level of uncertainty~(\textbf{larger embedding variance}, i.e., larger contours, and \textbf{closer to the decision boundary}). 
From the figure, we have the following observations:
\begin{itemize}[leftmargin=18pt]
    \item All three Bayesian approaches produce higher embedding variance for incorrect predictions and lower embedding variance for correct predictions. However, \ours achieves significantly larger embedding variance compared to the baselines, consistent with the quantitative evaluation shown in \Tabref{tab:main-llama}. \ours's produced variance is higher for the incorrect predictions, demonstrating its accurate uncertainty estimation even in the embedding space. 
    \item In \ours, the mean embedding produced by sampling weights from the approximate posterior is closer to the decision boundary than the embedding generated by the mean of weights (\starmark$\rightarrow$\trianglemark). This effect is most apparent when the prediction is incorrect, consistent with the quantitative results yielded from the final softmax layer of the model. Again, this demonstrates \ours's Bayesian inference can bring the final prediction closer to the ground truth. 
\end{itemize}



\newpage
\section*{NeurIPS Paper Checklist}

\begin{enumerate}

\item {\bf Claims}
    \item[] Question: Do the main claims made in the abstract and introduction accurately reflect the paper's contributions and scope?
    \item[] Answer: \answerYes{} 
    \item[] Justification: The abstract and introduction contains summarized claims about this paper that accurately reflect our contributions. 
    \item[] Guidelines:
    \begin{itemize}
        \item The answer NA means that the abstract and introduction do not include the claims made in the paper.
        \item The abstract and/or introduction should clearly state the claims made, including the contributions made in the paper and important assumptions and limitations. A No or NA answer to this question will not be perceived well by the reviewers. 
        \item The claims made should match theoretical and experimental results, and reflect how much the results can be expected to generalize to other settings. 
        \item It is fine to include aspirational goals as motivation as long as it is clear that these goals are not attained by the paper. 
    \end{itemize}

\item {\bf Limitations}
    \item[] Question: Does the paper discuss the limitations of the work performed by the authors?
    \item[] Answer: \answerYes{} 
    \item[] Justification: Our main theorems~(\Thmref{thm:posterior} and \ref{thm:prior}) state the hidden assumptions of the low-rank structure of the approximate posterior we propose for LLMs, and we have discussed the similar topics we do not consider in the Remark.
    \item[] Guidelines:
    \begin{itemize}
        \item The answer NA means that the paper has no limitation while the answer No means that the paper has limitations, but those are not discussed in the paper. 
        \item The authors are encouraged to create a separate "Limitations" section in their paper.
        \item The paper should point out any strong assumptions and how robust the results are to violations of these assumptions (e.g., independence assumptions, noiseless settings, model well-specification, asymptotic approximations only holding locally). The authors should reflect on how these assumptions might be violated in practice and what the implications would be.
        \item The authors should reflect on the scope of the claims made, e.g., if the approach was only tested on a few datasets or with a few runs. In general, empirical results often depend on implicit assumptions, which should be articulated.
        \item The authors should reflect on the factors that influence the performance of the approach. For example, a facial recognition algorithm may perform poorly when image resolution is low or images are taken in low lighting. Or a speech-to-text system might not be used reliably to provide closed captions for online lectures because it fails to handle technical jargon.
        \item The authors should discuss the computational efficiency of the proposed algorithms and how they scale with dataset size.
        \item If applicable, the authors should discuss possible limitations of their approach to address problems of privacy and fairness.
        \item While the authors might fear that complete honesty about limitations might be used by reviewers as grounds for rejection, a worse outcome might be that reviewers discover limitations that aren't acknowledged in the paper. The authors should use their best judgment and recognize that individual actions in favor of transparency play an important role in developing norms that preserve the integrity of the community. Reviewers will be specifically instructed to not penalize honesty concerning limitations.
    \end{itemize}

\item {\bf Theory Assumptions and Proofs}
    \item[] Question: For each theoretical result, does the paper provide the full set of assumptions and a complete (and correct) proof?
    \item[] Answer: \answerYes{} 
    \item[] Justification: In the main statement of the theorems and their proof, we clearly include the assumptions we make underlying them.
    \item[] Guidelines:
    \begin{itemize}
        \item The answer NA means that the paper does not include theoretical results. 
        \item All the theorems, formulas, and proofs in the paper should be numbered and cross-referenced.
        \item All assumptions should be clearly stated or referenced in the statement of any theorems.
        \item The proofs can either appear in the main paper or the supplemental material, but if they appear in the supplemental material, the authors are encouraged to provide a short proof sketch to provide intuition. 
        \item Inversely, any informal proof provided in the core of the paper should be complemented by formal proofs provided in appendix or supplemental material.
        \item Theorems and Lemmas that the proof relies upon should be properly referenced. 
    \end{itemize}

    \item {\bf Experimental Result Reproducibility}
    \item[] Question: Does the paper fully disclose all the information needed to reproduce the main experimental results of the paper to the extent that it affects the main claims and/or conclusions of the paper (regardless of whether the code and data are provided or not)?
    \item[] Answer: \answerYes{} 
    \item[] Justification: We have provided the full settings of our experiments in \Secref{sec:experiments} and \appref{app:implementation}. 
    \item[] Guidelines:
    \begin{itemize}
        \item The answer NA means that the paper does not include experiments.
        \item If the paper includes experiments, a No answer to this question will not be perceived well by the reviewers: Making the paper reproducible is important, regardless of whether the code and data are provided or not.
        \item If the contribution is a dataset and/or model, the authors should describe the steps taken to make their results reproducible or verifiable. 
        \item Depending on the contribution, reproducibility can be accomplished in various ways. For example, if the contribution is a novel architecture, describing the architecture fully might suffice, or if the contribution is a specific model and empirical evaluation, it may be necessary to either make it possible for others to replicate the model with the same dataset, or provide access to the model. In general. releasing code and data is often one good way to accomplish this, but reproducibility can also be provided via detailed instructions for how to replicate the results, access to a hosted model (e.g., in the case of a large language model), releasing of a model checkpoint, or other means that are appropriate to the research performed.
        \item While NeurIPS does not require releasing code, the conference does require all submissions to provide some reasonable avenue for reproducibility, which may depend on the nature of the contribution. For example
        \begin{enumerate}
            \item If the contribution is primarily a new algorithm, the paper should make it clear how to reproduce that algorithm.
            \item If the contribution is primarily a new model architecture, the paper should describe the architecture clearly and fully.
            \item If the contribution is a new model (e.g., a large language model), then there should either be a way to access this model for reproducing the results or a way to reproduce the model (e.g., with an open-source dataset or instructions for how to construct the dataset).
            \item We recognize that reproducibility may be tricky in some cases, in which case authors are welcome to describe the particular way they provide for reproducibility. In the case of closed-source models, it may be that access to the model is limited in some way (e.g., to registered users), but it should be possible for other researchers to have some path to reproducing or verifying the results.
        \end{enumerate}
    \end{itemize}

\item {\bf Open access to data and code}
    \item[] Question: Does the paper provide open access to the data and code, with sufficient instructions to faithfully reproduce the main experimental results, as described in supplemental material?
    \item[] Answer: \answerYes{} 
    \item[] Justification: we have submitted the anonymized code to Openreview, with the dependency specification of the packages and instructions for running the code.
    \item[] Guidelines:
    \begin{itemize}
        \item The answer NA means that paper does not include experiments requiring code.
        \item Please see the NeurIPS code and data submission guidelines (\url{https://nips.cc/public/guides/CodeSubmissionPolicy}) for more details.
        \item While we encourage the release of code and data, we understand that this might not be possible, so “No” is an acceptable answer. Papers cannot be rejected simply for not including code, unless this is central to the contribution (e.g., for a new open-source benchmark).
        \item The instructions should contain the exact command and environment needed to run to reproduce the results. See the NeurIPS code and data submission guidelines (\url{https://nips.cc/public/guides/CodeSubmissionPolicy}) for more details.
        \item The authors should provide instructions on data access and preparation, including how to access the raw data, preprocessed data, intermediate data, and generated data, etc.
        \item The authors should provide scripts to reproduce all experimental results for the new proposed method and baselines. If only a subset of experiments are reproducible, they should state which ones are omitted from the script and why.
        \item At submission time, to preserve anonymity, the authors should release anonymized versions (if applicable).
        \item Providing as much information as possible in supplemental material (appended to the paper) is recommended, but including URLs to data and code is permitted.
    \end{itemize}

\item {\bf Experimental Setting/Details}
    \item[] Question: Does the paper specify all the training and test details (e.g., data splits, hyperparameters, how they were chosen, type of optimizer, etc.) necessary to understand the results?
    \item[] Answer: \answerYes{} 
    \item[] Justification: We have provided the full settings of our experiments in \Secref{sec:experiments} and \appref{app:implementation}. 
    \item[] Guidelines:
    \begin{itemize}
        \item The answer NA means that the paper does not include experiments.
        \item The experimental setting should be presented in the core of the paper to a level of detail that is necessary to appreciate the results and make sense of them.
        \item The full details can be provided either with the code, in appendix, or as supplemental material.
    \end{itemize}

\item {\bf Experiment Statistical Significance}
    \item[] Question: Does the paper report error bars suitably and correctly defined or other appropriate information about the statistical significance of the experiments?
    \item[] Answer: \answerYes{} 
    \item[] Justification: Our experiments are repeated for 3 runs with different random seeds, and we have reported the standard deviation of all the methods in all the tables. 
    \item[] Guidelines:
    \begin{itemize}
        \item The answer NA means that the paper does not include experiments.
        \item The authors should answer "Yes" if the results are accompanied by error bars, confidence intervals, or statistical significance tests, at least for the experiments that support the main claims of the paper.
        \item The factors of variability that the error bars are capturing should be clearly stated (for example, train/test split, initialization, random drawing of some parameter, or overall run with given experimental conditions).
        \item The method for calculating the error bars should be explained (closed form formula, call to a library function, bootstrap, etc.)
        \item The assumptions made should be given (e.g., Normally distributed errors).
        \item It should be clear whether the error bar is the standard deviation or the standard error of the mean.
        \item It is OK to report 1-sigma error bars, but one should state it. The authors should preferably report a 2-sigma error bar than state that they have a 96\% CI, if the hypothesis of Normality of errors is not verified.
        \item For asymmetric distributions, the authors should be careful not to show in tables or figures symmetric error bars that would yield results that are out of range (e.g. negative error rates).
        \item If error bars are reported in tables or plots, The authors should explain in the text how they were calculated and reference the corresponding figures or tables in the text.
    \end{itemize}

\item {\bf Experiments Compute Resources}
    \item[] Question: For each experiment, does the paper provide sufficient information on the computer resources (type of compute workers, memory, time of execution) needed to reproduce the experiments?
    \item[] Answer: \answerYes{} 
    \item[] Justification: See \appref{sec:exp-mem cost}.
    \item[] Guidelines:
    \begin{itemize}
        \item The answer NA means that the paper does not include experiments.
        \item The paper should indicate the type of compute workers CPU or GPU, internal cluster, or cloud provider, including relevant memory and storage.
        \item The paper should provide the amount of compute required for each of the individual experimental runs as well as estimate the total compute. 
        \item The paper should disclose whether the full research project required more compute than the experiments reported in the paper (e.g., preliminary or failed experiments that didn't make it into the paper). 
    \end{itemize}
    
\item {\bf Code Of Ethics}
    \item[] Question: Does the research conducted in the paper conform, in every respect, with the NeurIPS Code of Ethics \url{https://neurips.cc/public/EthicsGuidelines}?
    \item[] Answer: \answerYes{} 
    \item[] Justification: We have read and fully considered the Code Of Ethics of NeurIPS, and we confirm we follow it strictly without any violation. 
    \item[] Guidelines:
    \begin{itemize}
        \item The answer NA means that the authors have not reviewed the NeurIPS Code of Ethics.
        \item If the authors answer No, they should explain the special circumstances that require a deviation from the Code of Ethics.
        \item The authors should make sure to preserve anonymity (e.g., if there is a special consideration due to laws or regulations in their jurisdiction).
    \end{itemize}

\item {\bf Broader Impacts}
    \item[] Question: Does the paper discuss both potential positive societal impacts and negative societal impacts of the work performed?
    \item[] Answer: \answerYes{} 
    \item[] Justification: We have some discussions on how our research can achieve a reliable LLM deployment with reduced harm to people by estimating the uncertainty of the prediction in \Secref{sec:intro}.
    \item[] Guidelines:
    \begin{itemize}
        \item The answer NA means that there is no societal impact of the work performed.
        \item If the authors answer NA or No, they should explain why their work has no societal impact or why the paper does not address societal impact.
        \item Examples of negative societal impacts include potential malicious or unintended uses (e.g., disinformation, generating fake profiles, surveillance), fairness considerations (e.g., deployment of technologies that could make decisions that unfairly impact specific groups), privacy considerations, and security considerations.
        \item The conference expects that many papers will be foundational research and not tied to particular applications, let alone deployments. However, if there is a direct path to any negative applications, the authors should point it out. For example, it is legitimate to point out that an improvement in the quality of generative models could be used to generate deepfakes for disinformation. On the other hand, it is not needed to point out that a generic algorithm for optimizing neural networks could enable people to train models that generate Deepfakes faster.
        \item The authors should consider possible harms that could arise when the technology is being used as intended and functioning correctly, harms that could arise when the technology is being used as intended but gives incorrect results, and harms following from (intentional or unintentional) misuse of the technology.
        \item If there are negative societal impacts, the authors could also discuss possible mitigation strategies (e.g., gated release of models, providing defenses in addition to attacks, mechanisms for monitoring misuse, mechanisms to monitor how a system learns from feedback over time, improving the efficiency and accessibility of ML).
    \end{itemize}
    
\item {\bf Safeguards}
    \item[] Question: Does the paper describe safeguards that have been put in place for responsible release of data or models that have a high risk for misuse (e.g., pretrained language models, image generators, or scraped datasets)?
    \item[] Answer: \answerNA{} 
    \item[] Justification: The paper does not pose any risks since there is no release of models.
    \item[] Guidelines:
    \begin{itemize}
        \item The answer NA means that the paper poses no such risks.
        \item Released models that have a high risk for misuse or dual-use should be released with necessary safeguards to allow for controlled use of the model, for example by requiring that users adhere to usage guidelines or restrictions to access the model or implementing safety filters. 
        \item Datasets that have been scraped from the Internet could pose safety risks. The authors should describe how they avoided releasing unsafe images.
        \item We recognize that providing effective safeguards is challenging, and many papers do not require this, but we encourage authors to take this into account and make a best faith effort.
    \end{itemize}

\item {\bf Licenses for existing assets}
    \item[] Question: Are the creators or original owners of assets (e.g., code, data, models), used in the paper, properly credited and are the license and terms of use explicitly mentioned and properly respected?
    \item[] Answer: \answerYes{} 
    \item[] Justification: we perform all the experiments under the license, and have cited work properly in the paper.
    \item[] Guidelines:
    \begin{itemize}
        \item The answer NA means that the paper does not use existing assets.
        \item The authors should cite the original paper that produced the code package or dataset.
        \item The authors should state which version of the asset is used and, if possible, include a URL.
        \item The name of the license (e.g., CC-BY 4.0) should be included for each asset.
        \item For scraped data from a particular source (e.g., website), the copyright and terms of service of that source should be provided.
        \item If assets are released, the license, copyright information, and terms of use in the package should be provided. For popular datasets, \url{paperswithcode.com/datasets} has curated licenses for some datasets. Their licensing guide can help determine the license of a dataset.
        \item For existing datasets that are re-packaged, both the original license and the license of the derived asset (if it has changed) should be provided.
        \item If this information is not available online, the authors are encouraged to reach out to the asset's creators.
    \end{itemize}

\item {\bf New Assets}
    \item[] Question: Are new assets introduced in the paper well documented and is the documentation provided alongside the assets?
    \item[] Answer: \answerNA{} 
    \item[] Justification: There is no new assets introduced in this paper.
    \item[] Guidelines:
    \begin{itemize}
        \item The answer NA means that the paper does not release new assets.
        \item Researchers should communicate the details of the dataset/code/model as part of their submissions via structured templates. This includes details about training, license, limitations, etc. 
        \item The paper should discuss whether and how consent was obtained from people whose asset is used.
        \item At submission time, remember to anonymize your assets (if applicable). You can either create an anonymized URL or include an anonymized zip file.
    \end{itemize}

\item {\bf Crowdsourcing and Research with Human Subjects}
    \item[] Question: For crowdsourcing experiments and research with human subjects, does the paper include the full text of instructions given to participants and screenshots, if applicable, as well as details about compensation (if any)? 
    \item[] Answer: \answerNA{} 
    \item[] Justification: No crowdsourcing experiments are involved in this paper.
    \item[] Guidelines:
    \begin{itemize}
        \item The answer NA means that the paper does not involve crowdsourcing nor research with human subjects.
        \item Including this information in the supplemental material is fine, but if the main contribution of the paper involves human subjects, then as much detail as possible should be included in the main paper. 
        \item According to the NeurIPS Code of Ethics, workers involved in data collection, curation, or other labor should be paid at least the minimum wage in the country of the data collector. 
    \end{itemize}

\item {\bf Institutional Review Board (IRB) Approvals or Equivalent for Research with Human Subjects}
    \item[] Question: Does the paper describe potential risks incurred by study participants, whether such risks were disclosed to the subjects, and whether Institutional Review Board (IRB) approvals (or an equivalent approval/review based on the requirements of your country or institution) were obtained?
    \item[] Answer: \answerNA{} 
    \item[] Justification: No IRB is involved in this paper.
    \item[] Guidelines:
    \begin{itemize}
        \item The answer NA means that the paper does not involve crowdsourcing nor research with human subjects.
        \item Depending on the country in which research is conducted, IRB approval (or equivalent) may be required for any human subjects research. If you obtained IRB approval, you should clearly state this in the paper. 
        \item We recognize that the procedures for this may vary significantly between institutions and locations, and we expect authors to adhere to the NeurIPS Code of Ethics and the guidelines for their institution. 
        \item For initial submissions, do not include any information that would break anonymity (if applicable), such as the institution conducting the review.
    \end{itemize}

\end{enumerate}

\end{document}